\documentclass[runningheads]{llncs}

\usepackage{eccv}

\usepackage{eccvabbrv}

\usepackage{graphicx}
\usepackage{booktabs}
\usepackage[dvipsnames]{xcolor}
\usepackage{pifont}
\usepackage[accsupp]{axessibility}  %

\usepackage{hyperref}
\newcommand{\xmark}{\color{Maroon}{\ding{55}}}
\newcommand{\mycheckmark}{\color{ForestGreen}{\ding{52}}}
\newcommand{\sr}{\texttt{SR}\xspace}
\newcommand{\spl}{\texttt{SPL}\xspace}
\newcommand{\nq}{\texttt{NQ}\xspace}
\newcommand{\fr}{\texttt{FR}\xspace}

\def\appendix{\textit{Appendix}\xspace}

\usepackage{orcidlink}

\usepackage{siunitx}

\usepackage[ruled,linesnumbered]{algorithm2e}
\usepackage{booktabs}
\usepackage{multirow}
\usepackage{multicol}
\usepackage{enumitem}
\SetKwComment{Comment}{/* }{ */}

\definecolor{phtalogreen}{HTML}{123524}
\definecolor{fgreen}{HTML}{228b22}
\definecolor{seablue}{HTML}{006994}
\definecolor{draftblue}{HTML}{116994}
\definecolor{algogreen}{HTML}{006400}
\definecolor{purpleblue}{HTML}{004194}
\definecolor{truepurple}{HTML}{800020}
\definecolor{mypurple}{HTML}{006400}
\definecolor{mypurple1}{HTML}{a43544}
\definecolor{lightred}{HTML}{F77189}
\definecolor{lightblue}{HTML}{39A7D0}
\definecolor{lightgreen}{HTML}{36ADA4}
\definecolor{grayish}{HTML}{b4c5dc}

\newcommand{\ourdataset}{QAsk-Nav-dataset\xspace}

\newcommand{\ourmethod}{Light-CoNav\xspace}
\newcommand{\ourmethodc}{Light-CoNav$^\circ$\xspace}
\newcommand{\ourqabenchmark}{QAsk-Nav\xspace}

\usepackage{listings}
\usepackage{xcolor}
\usepackage[many]{tcolorbox}

\lstdefinestyle{promptstyle}{
    basicstyle=\ttfamily\footnotesize,
    breaklines=true,
    showstringspaces=false,
    breakindent=0pt,
}

\newtcolorbox{beautifulprompt}{
    colback=gray!10,
    colframe=gray!50,
    boxrule=0.7pt,
    arc=4pt,
    left=2mm,
    right=2mm,
    top=1mm,
    bottom=1mm,
    breakable
}

\newcommand{\suppmat}{\textit{Supp. Mat.}\xspace}

\begin{document}

\title{Benchmarking Interaction, Beyond Policy: a Reproducible Benchmark for \\Collaborative Instance Object Navigation}

\titlerunning{Benchmarking Interaction, Beyond Policy}

\author{Edoardo Zorzi\inst{1,2} \and
Francesco Taioli\inst{3} \and
Yiming Wang\inst{4} \and
Marco Cristani\inst{2,5} \and
Alessandro Farinelli\inst{2} \and
Alberto Castellini\inst{2} \and
Loris Bazzani\inst{2}
}

\authorrunning{Zorzi, E. et al.}

\institute{Sapienza University of Rome, Rome, Italy \and
University of Verona, Verona, Italy \and
Polytechnic of Turin, \and
Fondazione Bruno Kessler\and
Reykjavik University}

\maketitle

\begin{abstract}
We propose \textit{Question-Asking Navigation} (\ourqabenchmark), the first reproducible benchmark for \textit{Collaborative Instance Object Navigation} (CoIN) that enables an explicit, separate assessment of embodied navigation and collaborative question asking.
CoIN tasks an embodied agent with reaching a target specified in free-form natural language under partial observability, using only egocentric visual observations and interactive natural-language dialogue with a human, where the dialogue can help to resolve ambiguity among visually similar object instances.
Existing CoIN benchmarks are primarily focused on navigation success and offer no support for consistent evaluation of collaborative interaction. 
To address this limitation, \ourqabenchmark provides (i) a lightweight \textit{question-asking protocol} scored independently of navigation, (ii) an enhanced \textit{navigation protocol} with realistic, diverse, high-quality target descriptions, and (iii) an open-source \textit{dataset}, that includes 28{,}000 quality-checked reasoning and question-asking traces for training, evaluation and analysis of interactive capabilities of CoIN models.
Using the proposed \ourqabenchmark benchmark, we develop \ourmethod, a lightweight unified model for collaborative navigation that is 3$\times$ smaller and 70$\times$ faster than existing modular methods, while outperforming state-of-the-art CoIN approaches in generalization to unseen objects and environments.
Project page at \href{https://benchmarking-interaction.github.io/}{https://benchmarking-interaction.github.io/}.
\end{abstract}

\section{Introduction}
\textit{Open-vocabulary Instance Object Navigation (ION)}~\cite{taioli2025coin, Liu_2025_WACV, khanna2024goatbench, ION} requires an embodied agent to navigate complex environments to reach object instances specified via human-defined free-form natural language instructions. 
In real-world scenes, where multiple visually and semantically similar object instances are present (\eg, multiple desk chairs \textit{v.s.} dining chairs), ION becomes particularly challenging as the agent needs to disambiguate the target instance from similar distractor instances~\cite{OVOV2024, PSL2024, khanna2024goatbench, VLFM2024}.
An emerging paradigm in ION shifts focus toward \textit{human-agent collaboration} \cite{taioli2025coin,han2025dialnav}, introducing Collaborative instance Object Navigation (CoIN). 
This task necessitates interactive reasoning and uncertainty evaluation to determine the optimal timing and content of questions to the human, to resolve environmental or linguistic ambiguity.

\begin{figure}[t!]
    \centering
    \includegraphics[width=1\linewidth]{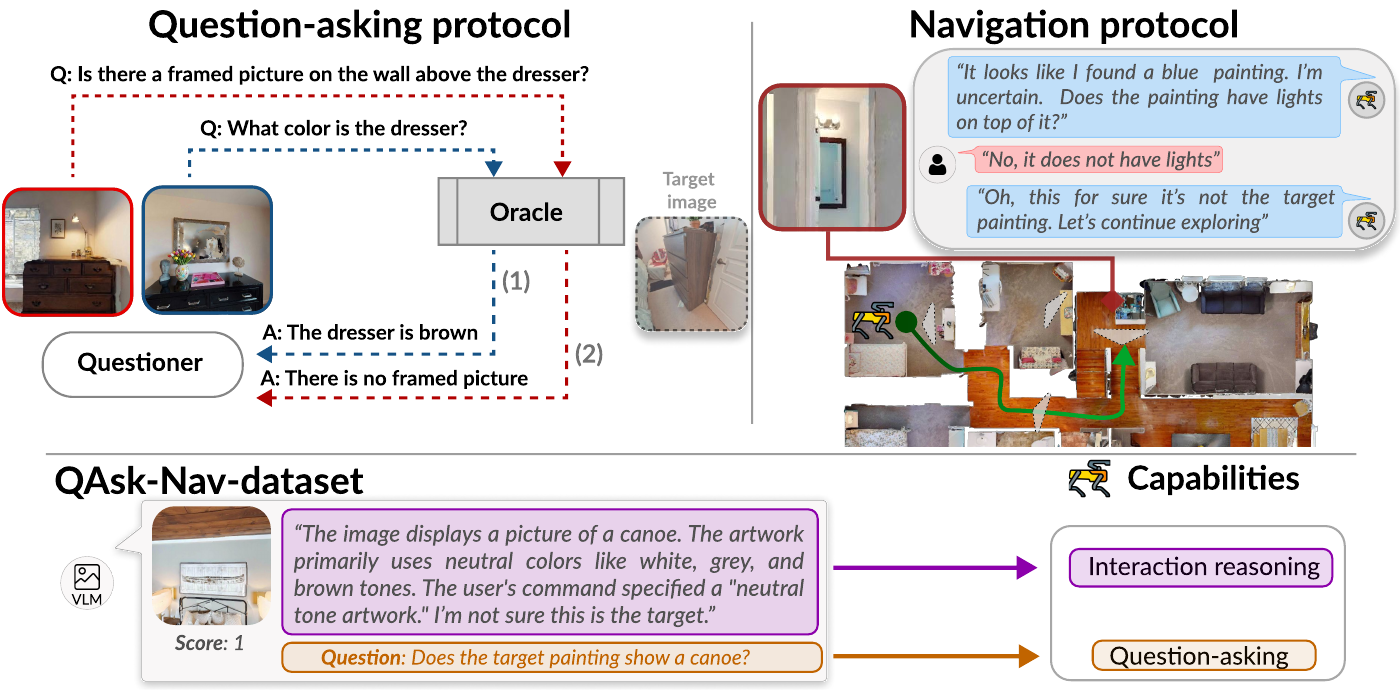}
    \caption{\ourqabenchmark introduces two distinct protocols for question asking (top-left) and for navigation (top-right), supported by a novel dataset with structured reasoning, and question annotations (bottom). Our dataset enables reproducible evaluation, training, and analysis under both protocols to study the agent's capabilities of interaction reasoning and question-asking.}
    \label{fig:teaser}
\end{figure}

Yet, the interactive nature of CoIN presents unique benchmarking challenges. 
The foundational CoIN benchmark~\cite{taioli2025coin} introduces additional collaborative settings into existing simulation frameworks. 
Human-agent interaction is evaluated within the navigation episodes, leading to four main limitations: {(i}) the performance is entangled with navigation policies, thus cannot isolate and measure solely the performance on human-agent interaction reasoning; {(ii}) the evaluation protocol is time-consuming as it requires the execution of the full navigation episode (even the most modern embodied simulators~\cite{ICCV_habitat, AI2Thor, Igibson} introduce significant latency with up to 5 minutes per episode); {(iii}) the benchmark is not reproducible because the human responses are simulated and can differ at each run; 
{(iv}) goal instructions generated via early-generation Vision-Language Models (VLMs) often exhibit insufficient semantic richness and diversity, limiting their utility for complex navigation tasks.

To address these limitations, we introduce \textit{Question-Asking Navigation} (\ourqabenchmark), the first navigation benchmark to support the development and evaluation of collaborative navigation agents in a \textit{disentangled}, \textit{lightweight} and \textit{reproducible} manner, for both navigation and question-asking capabilities.
As shown in \cref{fig:teaser}, \ourqabenchmark proposes three crucial elements: (i) %
a novel \textit{question-asking evaluation protocol} that is disentangled from the navigation episodes, enabling fast and reproducible evaluation of interaction reasoning capabilities; (ii) an improved \textit{navigation evaluation protocol} built on top of prior datasets \cite{taioli2025coin, khanna2024goatbench} with diverse, high-quality goal descriptions simulating realistic user commands, that greatly impacts the navigation performance of an embodied agent; (iii) a novel open-source \textit{dataset} (\ourdataset) featuring a collection of 28$,$000 high-quality reasoning and interaction traces generated by leveraging modern large VLMs~\cite{radford2021clip, Qwen3VL, Qwen25VL, InternVL} and text-guided image editing models~\cite{GEMINI, InstructPix2Pix} following the question-asking protocol. 
Building on this benchmark, we introduce a \textit{light-weight collaborative navigation agent} (\ourmethod), a small VLM finetuned with \ourdataset for interaction reasoning on the CoIN task. \ourmethod is substantially smaller (about 3$\times$ more compact), faster (by more than 70$\times$), and more interpretable (with much smaller reasoning traces by around 10$\times$) than prior state-of-the-art interaction models, while outperforming them both on navigation and question-asking tasks when evaluated on \ourqabenchmark.

In summary, our main contributions are the following:
\begin{enumerate}[noitemsep,topsep=0pt,leftmargin=*]
    \item We propose a novel {question-asking protocol} disentangled from navigation, enabling lightweight, reproducible evaluation.
    \item We enhance the {navigation protocol} with realistic, high-quality task descriptions that improve embodied agent navigation performance.
    \item We create an open-source \ourdataset with high-quality reasoning and question-asking traces to develop CoIN agents with interactive capability.
    \item We introduce \ourmethod, a lightweight model for {interaction reasoning} on the CoIN task, outperforming complex modular architectures.
   
\end{enumerate}

\section{Related Work}

Embodied agents have been evaluated on a variety of navigation tasks.
In \textit{object navigation}~\cite{THDA, OVOV2024, khanna2024goatbench, ZSON, anderson2018evaluation, batra2020objectnavevaluation, MULTION}, the agent is tasked with reaching any object belonging to a given category. 
In contrast, \textit{instance object navigation}~\cite{khanna2024goatbench, taioli2025coin, barsellotti24PIL, ION, Liu_2025_WACV}, requires the agent to reach a {specific} object instance described in natural language.
Similarly, \textit{instance image navigation}~\cite{PSL2024, krantz2023navigating, Krantz2022Instance, khanna2024goatbench} challenges the agent to locate a target specified via an image.
CoIN, as introduced in \cite{taioli2025coin}, extends these tasks by considering an agent that collaborates with the human.

\noindent \textbf{Benchmarks.}
Several benchmarks have been recently proposed for all these navigation tasks. 
OVON \cite{OVOV2024} uses realistic 3D scans from HM3D \cite{hm3d_sem} to build a large object navigation benchmark with the goal of fixing the poor generalization issues of models trained on synthetically generated scenes~\cite{HOMEROBOT}. GOAT-Bench~\cite{khanna2024goatbench} offers a comprehensive benchmark featuring three types of goal specifications: target images, object categories, and natural-language descriptions.
PIN \cite{barsellotti24PIL} and CoIN-Bench \cite{taioli2025coin} focuses on testing navigation towards specific instances of objects in photo-realistic scenarios~\cite{hm3d_sem}, including also distractor objects~\cite{AI2Thor, ICCV_habitat}. Among these, CoIN-Bench \cite{taioli2025coin} is the only benchmark that focuses on human-agent collaboration. 
However, prior works fail to decouple navigation from question-answering evaluation, and their reliance on simulator-heavy navigation episodes \cite{ICCV_habitat} introduces significant computational overhead.
\ourqabenchmark is the first to support the development and evaluation of collaborative navigation agents in a {disentangled}, {lightweight} and {reproducible} manner, for both navigation and question-asking capabilities.

\noindent \textbf{Interactive Navigation.}
Early work explored dialogue-guided navigation, where agents rely on human-provided guidance to complete tasks. For example,~\cite{alex_arena_nips, cvdn,rmm} introduce dialogue-based navigation benchmarks built from human-annotated interactions.
Many existing approaches restrict interaction to predefined formats~\cite{grounding_mate}, limited communication channels~\cite{majumdar2023findthis,framework_request_information}, or not free-form dialogue~\cite{singh2022ask4help,self_motivated,dialfred,Shen_2025_WACV}. 
Other works impose structural constraints on when or how interaction occurs: HANNA~\cite{hanna} allows interaction only at predefined locations, while
in~\cite{zhang-choi-2025-clarify} clarification decisions rely on manually annotated disambiguation intents. %
More recent work explores richer interaction settings: 
ZIPON~\cite{think_act_ask} studies personalized object retrieval, but the interaction is limited to ground-truth annotations.
DialNav~\cite{han2025dialnav} explores a setting in which a {navigator} reaches a goal region with assistance from a remote {guide}, who must infer the navigator's position.
AIUTA~\cite{taioli2025coin} introduces a free-form modular framework that focuses on human–agent interaction, combining VLMs and LLMs. However, AIUTA suffers from high latency due to the iterative procedure of its question-asking module, making real-world deployment challenging.
In this work, we show that the complexities of previous modular approaches can be reduced.
As a result, we introduce \ourmethod, a unified model that is substantially smaller, faster, and simpler than previous methods.

\section{The QAsk-NAV Benchmark}
In this section, we will first provide the definition for the CoIN task (\cref{subsec:task_definition}), followed by our proposed benchmark. Specifically,
\ourqabenchmark introduces three elements: 
(i) a question-asking evaluation protocol that is disentangled from the navigation episode (\cref{subsec:q-a}), (ii) an improved navigation evaluation protocol with high-quality task descriptions simulating realistic user commands (\cref{sec:navbench}), and (iii) a modern dataset (QAsk-NAV-dataset) featuring a collection of diverse and high-quality reasoning and interaction traces (\cref{subsec:ourdataset}).

\subsection{Task Definition} \label{subsec:task_definition}
In the CoIN task~\cite{taioli2025coin}, a navigation agent must collaborate with a user to reach a specific object instance expressed in free-form natural language, that can convey an arbitrary level of details, \eg, ``the bed'' or ``the unmade bed near the open window''.
The human-agent interaction in CoIN is essential as the agent needs to reason upon visual observations and the updated knowledge about the \textit{target} to ask user questions in order to effectively disambiguate the \textit{target} from other visually or semantically similar objects, \ie, \textit{distractors}.

Specifically, we consider the agent operating in a continuous environment. 
A CoIN navigation episode begins upon receiving a free-form natural language instruction describing the target $D$ provided by the user. We assume the target object is present in the environment.
At each time step $t$, the agent, parameterized by a policy $\pi$, receives an observation $O_t$ and selects an action $a_t \in A=\{$\texttt{Forward 0.25m, Turn Right 15°, Turn Left 15°, Stop, Ask}$\}$. 
Navigation terminates when the agent selects the \texttt{Stop} action or when the maximum number of actions is reached.
When the agent selects the \texttt{Ask} action, it generates a template-free, open-ended natural language question, denoted as $Q \in \mathcal{Q}$. The user’s corresponding free-form natural language response, denoted $R \in \mathcal{R}$, is stored in the response set, forming the interaction context $C$, which can be leveraged to enable the agent to ask relevant and disambiguating questions.

\subsection{Question-asking Protocol}\label{subsec:q-a}
\begin{figure}[t!]
    \centering
    \includegraphics[width=0.3\linewidth]{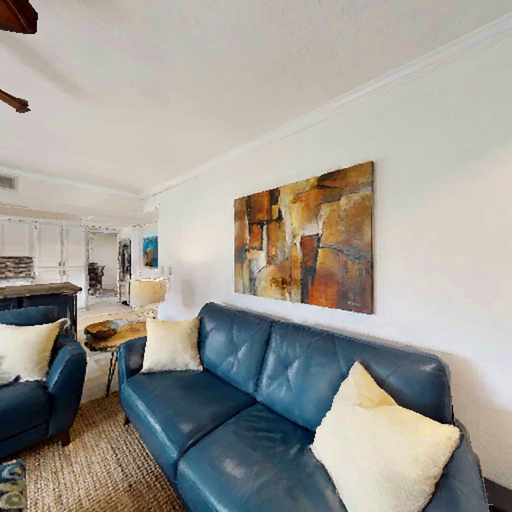}
    \includegraphics[width=0.3\linewidth]{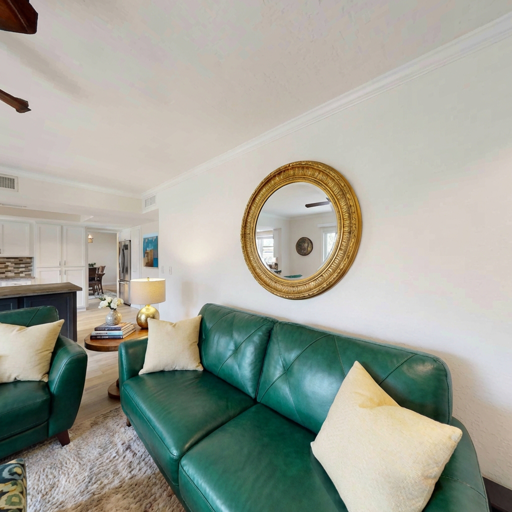}
    \includegraphics[width=0.3\linewidth]{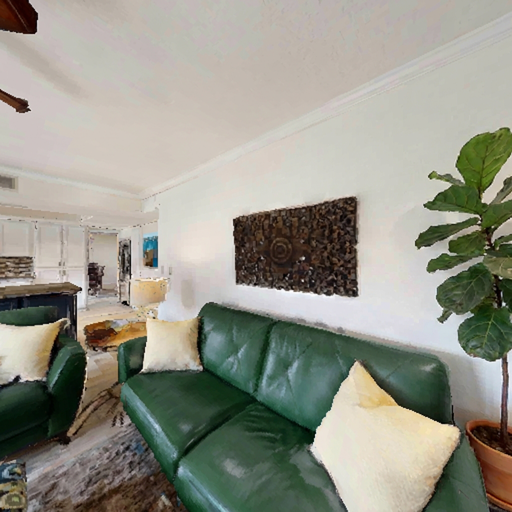}
    \caption{Examples from \ourqabenchmark. Left: original image. Center: distractor with an altered sofa and painting colors. Right: distractor with an altered plant and painting.}
    \label{fig:image_variations}
\end{figure}
To decouple the interactive question-asking property from a CoIN navigation episode, we design a reinforcement learning-style \cite{sutton2020rl} protocol (\cref{fig:teaser}, top-left) that focuses solely on human-agent interaction inspired by~\cite{guesswhat2017}. 
The goal is to evaluate the agent's capability in asking clarifying questions about a target object when observing a candidate instance, given the initial description 
$D$, without running a full CoIN episode.
Our protocol aims to mimic the situation during a CoIN episode where the agent faces a candidate object and needs to ask a question to the user. Specifically, we propose to have a \textit{questioner} role representing the navigation agent who asks questions, and an \textit{oracle} role, simulating the user answering the agents' questions (see \cref{fig:teaser} top-left). 
The oracle knows how the target object looks like since it has access to the image of the goal (not available to the questioner). 
The questioner is provided with the natural-language description $D$ of the target object. 
To test multi-turn question-asking behavior for visual disambiguation, the protocol provides multiple images of distractors common in real-world scenes.
The distractors are presented to the questioner at each turn, who must determine whether they correspond to the target, or ask a clarifying question to the oracle. 
In the following, we describe in detail how the different components of the protocol are used to obtain a realistic benchmark.

\noindent\textbf{Distractor Synthesis.} 
A naive approach to generating distractors is to directly source observations from navigation episodes.
However, due to the limited diversity of present instances, such an approach does not create challenging observations of distractors.
We therefore opt to a generative way to generate diverse and challenging distractors exploiting advances in text-guided image editing model, \ie, \texttt{Gemini-3-pro-image} \cite{deepmind_gemini3proimage}. 
We source images from existing benchmarks, CoIN-Bench~\cite{taioli2025coin} and GOAT-Bench~\cite{khanna2024goatbench}, and perform image editing of fine-grained attributes of the objects that are essential to uniquely identify an instance, including \textit{color}, \textit{salient feature}, and \textit{contextual} elements. For each semantic category, we select 10 base images and obtain 5 variants of each by modifying these attributes, creating sets of images that differ from the original image in fine-grained details.
This creates challenging, yet realistic scenarios where the distractors are very similar to the target (see \cref{fig:image_variations}).

\noindent\textbf{Target Descriptions.} 
In addition to distractors, we synthetically generate the descriptions of the target object instances $D$. Different from prior benchmark~\cite{taioli2025coin}, which only provide a brief description $D$, we aim to facilitate the study on how the specificity of the description influences the question-asking capabilities of the questioners and the navigation difficulty.
To this end, we create six levels of descriptions with increasing specificity, ranging from minimal category-level cues, to highly detailed descriptions: (i) \textit{category} (\eg, `bed'); (ii) \textit{category and color} (\eg, `blue and dark brown bed'); (iii) \textit{color-feature} which includes category, color, and a salient feature such as texture of the object (\eg, `blue patchwork bed'); (iv) \textit{context}, which includes category plus surrounding context (\eg, `bed next to a wooden nightstand'); (v) \textit{color-context}, that includes category, color, and context (\eg, `blue and dark brown bed next to a wooden nightstand'); (vi) \textit{color-context-feature}, that includes category with color, context, and feature (\eg, `blue patchwork bed near a wooden nightstand'). 
We provide qualitative examples in \cref{sec:appendix_qask} of the \suppmat 

\begin{figure}[tp]
    \centering
    \includegraphics[width=1\linewidth]{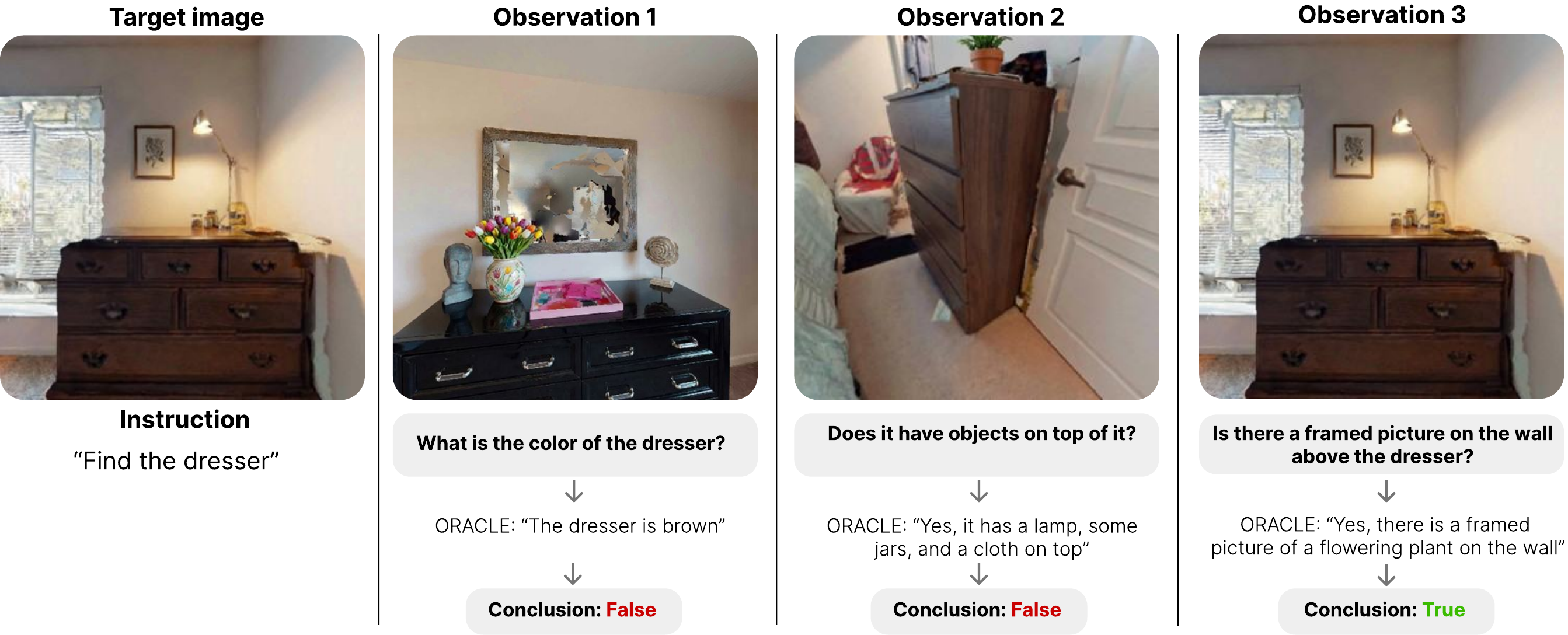}
    \caption{Episode from \ourqabenchmark. Column 1: the target image available only to the oracle and the navigation instructions provided to the agent. 
    Columns 2-3: different observations, including image, question, and answer from the oracle. 
    Before each question and conclusion, the model produces a reasoning trace, motivating the decision. More detailed examples in \suppmat \cref{sec:appendix_qask}.}
    \label{fig:qask_nav_episode}
\end{figure}

\noindent\textbf{Interactive Episode.} 
Our protocol incorporates multi-turn episodes, featuring the persistent interaction required for effective human-agent collaboration in real-world scenarios.
We construct the interactive episodes with the generated diverse distractor images and instructions~$D$ of multi-level specificity. Specifically, each episode incorporates the following data:
(i) a target image, which represents the target instance; (ii) a natural-language description of the target; (iii) a sequence of distractors of the same category. In total, we collect data for $\sim$1400 episodes, which we split into a training and test set, comprising, respectively, data for $\sim$1,000 and $\sim$400 episodes (details in \suppmat \cref{sec:appendix_qask}).
As shown in \cref{fig:qask_nav_episode}, for each episode, the oracle is given access to the image of the target object instance, while the questioner has only access to the description $D$. 
We configure the episode to show distractors consecutively and finish with the target instance. 
At each step, the questioner is presented with an observation image, and it may either ask a question about the target or output a binary decision indicating whether the current observation matches the target.
The episode continues if the agent correctly decides if the observation belongs to a distractor; otherwise, the episode terminates.
Once the questioner outputs a decision for the final observation, the episode also terminates. 
As the episode always starts with distractors followed by the target, the agent may shortcut the task by always returning a negative decision. To mitigate, we vary the number of observations across episodes with more than 4 candidates on average. We also measure the finish rate, which measures the ratio of completed episodes (requiring all correct decisions) (\cref{sec:results}). The shortcut strategy would result in a finish rate of 0.

\subsection{Navigation Protocol}\label{sec:navbench}
To evaluate the navigation capabilities of different models (\cref{fig:teaser}, top-right), we propose a navigation protocol based on CoIN-Bench \cite{taioli2025coin}. CoIN-Bench comprises $\sim$1,600 navigation episodes, picked from GOAT-Bench~\cite{khanna2024goatbench}, based on the presence of distractor objects. 
As in GOAT-Bench, CoIN-Bench uses real-world 3D scene scans~\cite{hm3d_sem} within the Habitat simulator~\cite{ICCV_habitat}. 

\begin{figure}
    \centering
    \includegraphics[width=1\linewidth]{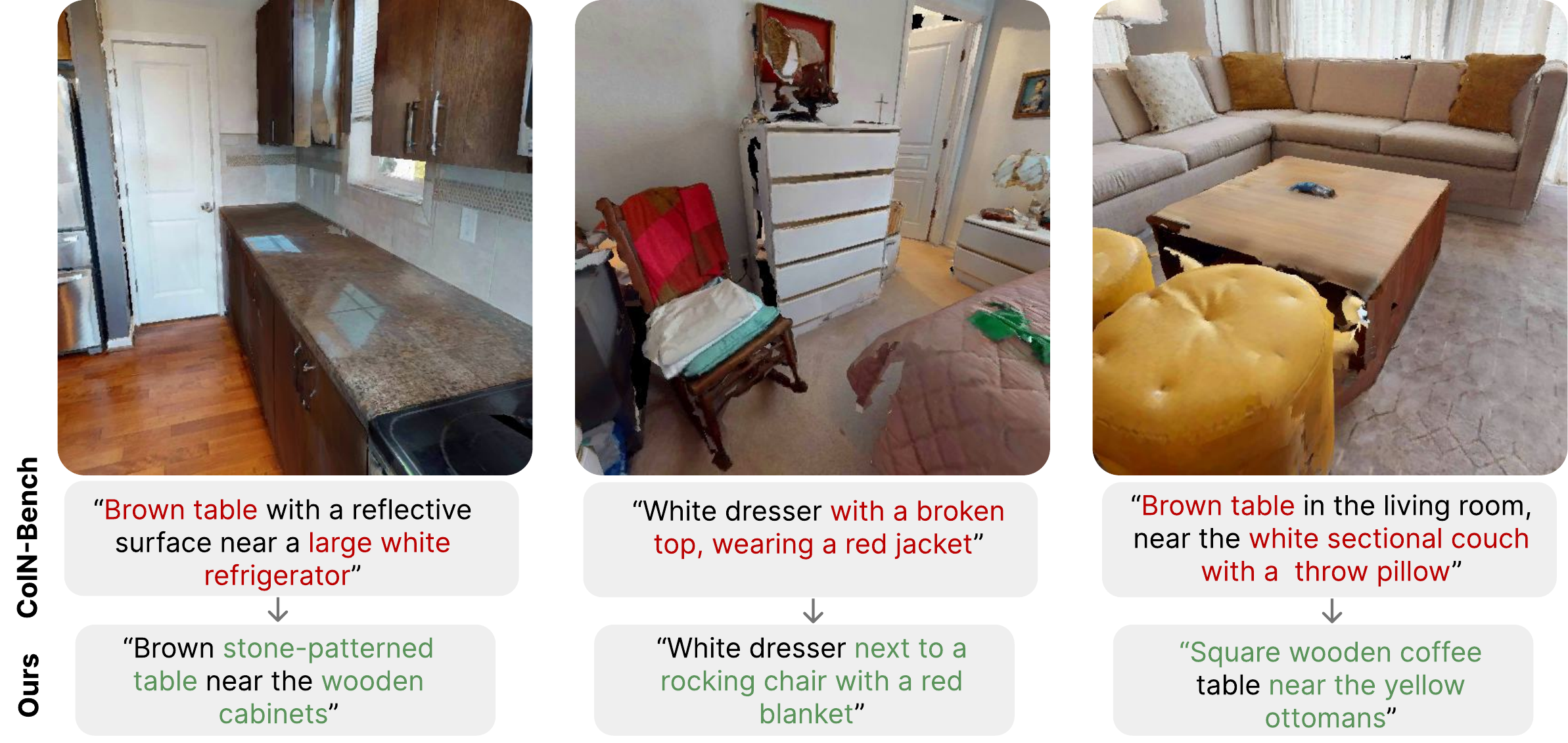}
    \caption{\textbf{Updated Task Annotations}. Comparison between CoIN-Bench \cite{taioli2025coin} annotations and our \ourqabenchmark annotations. Old annotations often contain confusing, nonsensical, or hallucinated descriptions, while new annotations are clearer and more accurate.}
    \label{fig:annotations}
\end{figure}

CoIN requires task instructions, \ie, the description of an instance object to initiate the navigation episodes. However, constrained by the adopted VLM (\texttt{LLaVA 1.6 Mistral 7B}~\cite{Liu2024Llavanext}), the quality of the instance description in CoIN-Bench is of limited quality, with wrong or hallucinated attributes (see \cref{fig:annotations}).
Moreover, to simulate the human user that provides responses to the navigation agent, CoIN-Bench leverages the same VLM~\texttt{LLaVA 1.6 Mistral 7B} as oracle. Thus, the interaction quality is also limited.

We overcome these limitations by leveraging a large VLM (\texttt{Gemini-3-pro}~\cite{deepmind_gemini3pro}) to generate high-quality descriptions for each target object, and use a more advanced and larger VLM  (\texttt{Qwen3-VL-30B-A3B} \cite{Qwen3VL}) as an oracle that is integrated within the Habitat simulator~\cite{ICCV_habitat}. In the following, we describe in detail the data split, and the target descriptions, and the episode execution during evaluation.

\noindent\textbf{Data split.} We follow the evaluation splits in prior works~\cite{khanna2024goatbench, taioli2025coin}: \textit{Val Seen}, \textit{Val Seen Synonyms}, and \textit{Val Unseen}. \textit{Val Seen} contains the same objects and categories as the \textit{Train} split, but different episodes (starting locations and distractors), the \textit{Val Seen Synonyms} split uses the same object categories but different descriptions (using synonyms), to test adaptation to lexical variations, whereas the \textit{Unseen} split, instead, contains completely novel objects. They comprise, respectively, 831, 359, and 459 episodes, for a total of 1,649 episodes (more information in the \suppmat, \cref{sec:appendix_coin}).

\noindent\textbf{Target Descriptions.} 
We leverage commercial tools such as \texttt{Gemini-3-pro} \cite{deepmind_gemini3pro} to ensure a faithful and complete description of each target object.
We ensure the description quality with manual inspection using a subsampled set of descriptions, where human annotators are tasked to choose preferred, and to note hallucinations and imprecise details.
Additionally, we evaluate on the full set using a large language model (LLM) as a judge~\cite{judging_llm_arena, g_eval, kim2024prometheus}. Our descriptions are preferred in more than 80\% of the cases, with very few inaccurate details (further information in \cref{sec:appendix_annotations} of the \suppmat).  This confirms that our descriptions are better in terms of precision, clarity with reduced hallucinations. On average, our descriptions are shorter than those in CoIN-Bench: $9.66$ words (std $1.81$) compared to $17.53$ words (std $9.46$) in CoIN-Bench. Note that we use the type \textit{(vi)} task descriptions in the navigation protocol (see \cref{subsec:q-a}).

\noindent\textbf{Navigation Episode.} The navigation agent starts from a random position in the environment, and is provided with a task description $D$ of a target. The oracle is initialized with access to the target image. The agent navigates in the environment, taking, at each time-step $t$, an action $a_t$. When $a_t = \texttt{Ask}$, the question is fed to the oracle, which faithfully responds to it using the question $Q$ contained in $a_t$, and the target image. A navigation is successful if the agent finds the target instance within the maximum number of steps.

\subsection{\ourdataset}\label{subsec:ourdataset}
Due to the interactive nature in CoIN, it is challenging to create dataset of sufficient scale to support model training to enhance the interaction reasoning in such collaborative setting. In fact, prior work only addressed interaction reasoning at inference time. To bridge this critical gap, we propose \ourdataset (\cref{fig:teaser}, bottom), the first dataset designed for training CoIN models to improve their reasoning and question-asking capabilities. 
The dataset is built by combining our protocols with modern generative models~\cite{GEMINI, GEMINI25, deepmind_gemini3proimage, Qwen3VL, InternVL, Qwen25VL}. 

\begin{figure}[t]
    \centering
    \includegraphics[width=0.97\linewidth]{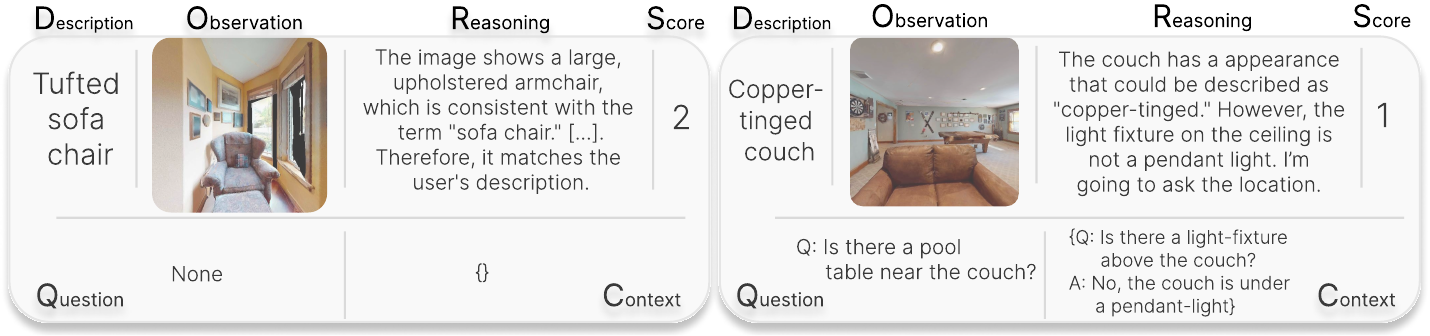}
    \caption{Two samples $(D,O,R,S,Q,C)$ from \ourdataset. In the first example, the description $D$ matches the observation $O$ and thus the reasoning $R$ and score $S$ indicate a match (between $D$ and $O$). In the second example, the description is ambiguous, and previous answers did not help: therefore $R$ and $S$ express uncertainty.
    }
    \label{fig:example_rqion}
\end{figure}

\ourdataset contains high-quality reasoning traces together with their uncertainty scores and questions.
Each sample of \ourdataset is a tuple $(D, O, R, S, Q, C)$. $D$ is a description of the target object instance of category $c$. $O$ is a visual observation of an object, featuring either distractors or the target. $R$ is a reasoning trace based on $O$ and $D$. $S \in \{0,1,2\}$ is an uncertainty score, indicating the degree of the match of $O$ with $D$, as motivated by the reasoning $R$. $S=0$ means that the model is certain that $O$ is not the target object, $S=1$ means the model is unsure about the match, and $S=2$ means that the model is certain that the observed object corresponds to the target object. $Q$ is a question if $S=1$, None otherwise. $C$ is the context containing previous question-answer pairs that can be used to inform what to ask. 
In total, it comprises approximately 28,000 reasoning traces, split into three partitions (following the standard protocol adopted by other benchmarks \cite{khanna2024goatbench, taioli2025coin}): \textit{Train} (15.98k), \textit{Val seen} (6.38k), and \textit{Val unseen} (5.03k).
See \cref{fig:example_rqion} for two examples.

We generate these samples starting from images $O$ collected from our navigation protocol (target objects and distractor images), and from other navigation benchmarks, such as GOAT-Bench~\cite{khanna2024goatbench}, and a mix of manually-curated and generated descriptions $D$. 
For each $(D,O)$ pair, we prompt a commercial LLM (\texttt{Gemini-3-pro}~\cite{deepmind_gemini3pro}) to determine whether the description $D$ matches with the observation $O$, that is, whether $O$ is an observation of the target object described by $D$. We prompt it to generate a reasoning $R$ and an uncertainty score $S$, plus a question $Q$ if unsure. The final sample is created by merging manually verified $(R, S, Q)$ and $(D,O,C)$ with $C = \emptyset$.
To increase sample diversity, we generate additional samples by running a large VLM as a questioner, by adopting our question-asking protocol, obtaining multiple reasoning traces and questions. From these rollouts (using the episode data from the training set), $D$ is the starting description and $O$ is the image distractor, while the reasoning $R$, the uncertainty score $S$ and the question $Q$ are obtained from the questioner's decision (the binary decisions, ``not match'' and ``match'', correspond to scores 0 and 2, respectively, and a question corresponds to a score of $1$). 
When the navigation agent asks a question for the first time, the context is empty $C=\emptyset$, otherwise, the previous questions and answers are added to $C$.

\section{\ourmethod: A Light Collaborative Navigation Agent}\label{sec:method}

Prior CoIN methods~\cite{taioli2025coin, han2025dialnav} implement collaboration through multi-stage modular pipelines that combine multiple LLM and VLM components. In commonly used evaluations, interaction quality is mainly reflected through end-to-end navigation outcomes, making it difficult to isolate the contribution of each interaction module and often encouraging added complexity. Our benchmark, instead, provides separate protocols and metrics for navigation and question asking. Together with \ourdataset, it supports training a compact model and shows that strong collaborative performance can be achieved without complex modular architectures.
We proved this by designing our navigation agent, called \ourmethod, that is finetuned for improved reasoning and question-asking capabilities on \ourdataset. %

\noindent\textbf{Architecture and Navigation.} 
\ourmethod consists of a base VLM architecture composed of a single, lightweight, multimodal transformer $\phi$ \cite{vaswani2017transformer}, derived from \texttt{Qwen2.5-VL-7B-Instruct} \cite{Qwen25VL}, which layers on top of an existing navigation policy $\pi$\footnote{\ourmethod deals only with human collaboration (i.e., question asking), hence it can be used with many navigation policies. The only requirement to use it is that the policy is modular and employs an object detector. Here we use VLFM \cite{VLFM2024}.}. 
The navigation policy $\pi$ explores and navigates the environment, conditioned on the initial task command $D$. Once it observes an image $O$ of an object belonging to the same category as the target object described by $D$, it passes the image to the VLM, which will output a {uncertainty score} $S$. %
Formally, $(R, S, Q) = \phi(p(D, O, C))$, with $p$ the prompt that takes in input $D$, $O$ and the context $C$ (containing the previous questions and answers, empty at the start), and returns a textual command to obtain the output reasoning $R$, uncertainty score $S$ and question $Q$. 
If the score is 2, it assigns this detection to the set of candidate objects, and $\pi$ starts navigating towards it; if the score is 0, it discards this observation and $\pi$ restarts the navigation; if the score is 1, it asks one or more questions $Q$ to the user, possibly requesting more details and clarifications. In case of multiple questions, the questions and corresponding answers are added to the context $C$ and passed to the model $\phi$ for the next interaction.

\noindent\textbf{Training.} We propose a two-stage training recipe to equip \ourmethod with reasoning and question-answering capabilities. %
In the first stage, the base model is fine-tuned via \textit{supervised fine-tuning} using \textit{reasoning} samples from \ourdataset without questions and empty context $C$.
This improves the model's ability to follow the given template and to produce high-quality reasoning traces and coherent uncertainty scores. Our tests showed that producing only a score $S$ (without a reasoning trace $R$) decreases model performance. %
After this first learning phase, the model is capable of producing a reasoning and uncertainty score, \emph{i.e.}, $(R, S, \emptyset) = \phi(p(D, O, \emptyset))$. 

At this stage, the model has not yet been trained to generate questions. In the second learning stage, we finetune it to acquire \textit{question-asking} capabilities. This is achieved by training on samples $(D,O,R,S,Q,C)$ that include a question, interleaved with a small number of samples without questions (with a ratio of 10:1), to avoid excessively perturbing the model's output.  In this phase, the model is also trained with non-empty contexts $C$.
After this phase, the model is capable of producing a reasoning and uncertainty score with questions, \emph{i.e.}, $(R, S, Q) = \phi(p(D, O, C))$. 
Further details in \cref{sec:appendix_training} of the \suppmat

\section{Experimental Results}\label{sec:results}

We conducted a thorough evaluation of the question-asking capabilities of various CoIN methods, following the protocol and episodes detailed in \cref{subsec:q-a}. 
Next, we assessed the navigation efficacy of the models using the evaluation framework defined in \cref{sec:navbench}.
The results are discussed in \cref{subsec:q-a-results} and \cref{subsec:navbench-results}.

\subsection{Results on Question-Asking}\label{subsec:q-a-results}

\textbf{Baselines.} 
To isolate the question-asking capabilities of CoIN methods, we compare the interactive module of AIUTA~\cite{taioli2025coin}\footnote{To ensure reproducible offline evaluation, we replaced the original online LLM in the AIUTA self-questioner module with \texttt{Qwen3-13B}~\cite{Qwen3VL}, providing similar performance with a total of 20B parameters.} and \ourmethod, which we mark as AIUTA$^\circ$ and \ourmethodc, respectively. 
In addition, we compare with the base \texttt{Qwen2.5-VL-7B}~\cite{Qwen25VL} to assess the impact of our fine-tuning and check the impact of the same number of parameters.
Finally, we establish performance upper bounds using two significantly larger models, \texttt{Qwen3-30B-VL-A3B}~\cite{Qwen3VL} and \texttt{Gemini-3-flash}~\cite{deepmind_gemini3flash_2025}, showing the capabilities of high-resource systems, typically inaccessible to navigation agents. 
Prompts are provided in the \suppmat

\noindent\textbf{Metrics.}
We evaluate performance using: Success Rate (\sr$\uparrow$), defined as the fraction of correct decisions per episode; Finish Rate (\fr $\uparrow$), the proportion of trajectories completed with all decisions correct; Number of Questions (\nq$ \downarrow$), the average number of questions asked per observation; and Execution Time (\texttt{Time} $\downarrow$), the average time in seconds required to complete an episode.
To mitigate potential biases inherent in individual VLMs, we employ a cross-model evaluation strategy using three distinct oracles:
\texttt{Gemini-2.5-pro} \cite{GEMINI25}, \texttt{Gemini-3-flash} \cite{GEMINI}, and \texttt{Qwen3-VL-30B-A3B-Instruct} \cite{Qwen3VL}. 
Reported metrics represent the mean performance aggregated across these three oracles. We use episode data from the test set.

\noindent\textbf{Results.}
\cref{tab:results_qask} reports the comparison of different methods using the question-answering protocol of \ourqabenchmark. 
We distinguish between navigation-ready models, those with the restricted parameters and memory efficiency required for robotic applications, and larger-scale, non-navigation-ready models.
\ourmethodc achieves a substantial improvement in \sr, \fr, and \nq compared to navigation-ready models, while reducing runtime. 
Notably, \ourmethodc significantly outperforms both the base \texttt{Qwen2.5-VL-7B}~\cite{Qwen25VL} and state-of-the-art AIUTA$^\circ$. 
This outcome underscores the effectiveness of our specialized training data. 
Despite being approximately 3$\times$ smaller in parameters, \ourmethodc yields superior question-answering performance compared to the significantly larger AIUTA$^\circ$. 
These results demonstrate that targeted supervision on \ourdataset effectively compensates for reduced model scale.
When comparing our method with non-navigation-ready models of \cref{tab:results_qask}, we notice that \ourmethodc outperforms \texttt{Qwen3-30B-VL-A3B} despite having 4$\times$ less parameters. 
While \ourmethodc achieves a \sr approximately $50\%$ of that attained by \texttt{Gemini-3-flash} \cite{deepmind_gemini3flash_2025}, this performance gap highlights the potential for further scaling. 
Future improvements could leverage our dataset synthesis recipe to generate even more specialized training data, alongside architectural refinements, to bridge the margin between navigation-ready models and high-resource frontier models.

\begin{table}[t]
    \caption{Performance comparison of methods on the \ourqabenchmark benchmark.
    The top section shows navigation-ready models (parameter size under 20B), while the bottom section includes larger or closed-source baselines.
    }
\small
\centering
    \begin{tabular}{lcccccccc}

    \toprule
    \textbf{Method} & Navigation ready & \textbf{SR} $\uparrow$ && \textbf{FR} $\uparrow$ && \textbf{NQ} $\downarrow$ && \textbf{Time} $\downarrow$ \\
    \midrule
    Qwen2.5-VL-7B \cite{Qwen25VL}   &  \mycheckmark & 0.0752 && 0.0034 && 2.7405 && 1.70  \\ 
    AIUTA$^\circ$ \cite{taioli2025coin} &  \mycheckmark & 0.3027 && 0.1993 && {1.1473} &&  177.83 \\
    \colorbox{grayish}{\ourmethodc} (ours) &  \mycheckmark &\textbf{0.4510} && \textbf{0.3317} && \textbf{0.1990} && \textbf{1.15}  \\\hline
    Qwen3-VL-30B-A3B  \cite{Qwen3VL}  & \xmark & 0.3205 && 0.2755 && 0.1174 && 1.58  \\
    Gemini-3-flash \cite{deepmind_gemini3flash_2025}  & \xmark & 0.9329 && 0.8943 && 0.9564 && 13.20  \\
    \bottomrule
    \end{tabular}
\label{tab:results_qask}
\end{table}

\begin{table*}[t]
\caption{Success Rate (\texttt{SR} $\uparrow$) and Finish Rate (\texttt{FR} $\uparrow$) across different annotation types of \ourqabenchmark for selected methods.}
\centering
\scriptsize
\begin{tabular}{lcc|cc|cc|rc|cc|cc}
\toprule
& \multicolumn{2}{c|}{Cat} 
& \multicolumn{2}{c|}{CatCol} 
& \multicolumn{2}{c|}{ColFeat} 
& \multicolumn{2}{c|}{Ctx} 
& \multicolumn{2}{c|}{ColCtx} 
& \multicolumn{2}{c}{ColCtxFeat} \\
\cmidrule(lr){2-3}
\cmidrule(lr){4-5}
\cmidrule(lr){6-7}
\cmidrule(lr){8-9}
\cmidrule(lr){10-11}
\cmidrule(lr){12-13}
\textbf{Method} 
& \textbf{SR}$\uparrow$ & \textbf{FR}$\uparrow$
& \textbf{SR}$\uparrow$ & \textbf{FR}$\uparrow$
& \textbf{SR}$\uparrow$ & \textbf{FR}$\uparrow$
& \textbf{SR}$\uparrow$ & \textbf{FR}$\uparrow$
& \textbf{SR}$\uparrow$ & \textbf{FR}$\uparrow$
& \textbf{SR}$\uparrow$ & \textbf{FR}$\uparrow$ \\
\midrule
AIUTA$^\circ$ \cite{taioli2025coin}
& \textbf{0.42} & \textbf{0.28} 
& 0.34 & 0.22 
& 0.30 & 0.19 
& 0.32 & 0.22 
& 0.24 & 0.15 
& 0.20 & 0.13 \\

\colorbox{grayish}{\ourmethodc} (ours)
&0.06 & 0.03
& \textbf{0.45} & \textbf{0.31}
& \textbf{0.58} & \textbf{0.44}
& \textbf{0.43} & \textbf{0.32}
& \textbf{0.61} & \textbf{0.42}
& \textbf{0.68} & \textbf{0.54}\\
\midrule
Qwen3-VL-30B-A3B \cite{Qwen3VL}
&0.02 & 0.01
& 0.24 & 0.18
& 0.40 & 0.34
& 0.34 & 0.29
& 0.47 & 0.41
& 0.54 & 0.47\\
\bottomrule
\end{tabular}

\label{tab:qask_results2}
\end{table*}

Next, we compare selected models on the six description types defined in the question-asking protocol (Section~\ref{subsec:q-a}) in order to further analyze their behavior with increasing difficulty of navigation instructions. 
As shown in \cref{tab:qask_results2}, \ourmethodc consistently outperforms AIUTA$^\circ$ and \texttt{Qwen3-VL-30B-A3B} in all settings except for the \textit{category} type, which is the least informative and most ambiguous description type. 
In the category case, \ourmethodc and \texttt{Qwen3-VL-30B-A3B} tend immediate action, prioritizing direct decisions over proactive questioning. 
This behavior indicates that, under ambiguous category-level descriptions, models within this parameter range ($\le 30$B) tend to favor execution over information-seeking. 
These results suggest that current navigation-ready architectures have yet to fully optimize the threshold for uncertainty-driven communication, marking a clear opportunity for future work in robust ambiguity resolution.
In all other description types, \ourmethodc achieves substantially higher performance. This is particularly evident for the annotation color-context-feature, which most closely matches the annotations used in the navigation episodes. 
We also observe that \ourmethodc performance increases monotonically with the amount of descriptive information (i.e., moving from the left to the right in \cref{tab:qask_results2}), reaching its best results in the most detailed setting. 
In contrast, AIUTA$^\circ$ does not consistently benefit from richer descriptions and, in some cases, degrades as more detail is introduced. 
These results highlight the stronger visual-textual grounding and the more effective use of contextual information of \ourmethodc.
Moreover, we also validate the oracle by comparing the performance with human responses, achieving similar performance. This validates that the oracle is a reliable proxy to evaluate agent's question-asking capabilities to human (more details in the \suppmat \cref{sec:appendix_qask}).

\subsection{Results on Navigation} \label{subsec:navbench-results}
\begin{table*}[t]
\caption{Results of the navigation task on the CoIN benchmark with our \ourqabenchmark annotations. %
}
\centering
\resizebox{0.95\textwidth}{!}{
\begin{tabular}{rccc cccc cccc|c cccc}
    \toprule
    \multirow{2}{*}{\textbf{Method}} & Features & \multicolumn{4}{c}{\textit{Val Seen}}  & \multicolumn{4}{c}{\textit{Val Seen Synonyms}}  & \multicolumn{1}{c}{} & \multicolumn{4}{c}{\textit{Val Unseen}}  \\ 
    \cmidrule{2-2} \cmidrule{4-6} \cmidrule{8-10} \cmidrule{13-15} 
         & Interaction && \scriptsize\textbf{SR}~$\uparrow$ & \scriptsize\textbf{SPL}~$\uparrow$ &\scriptsize\textbf{NQ}~$\downarrow$  && \scriptsize\textbf{SR}~$\uparrow$ & \scriptsize\textbf{SPL}~$\uparrow$ &\scriptsize\textbf{NQ}~$\downarrow$ &&& \scriptsize\textbf{SR}~$\uparrow$ & \scriptsize\textbf{SPL}~$\uparrow$ &\scriptsize\textbf{NQ}~$\downarrow$ \\ \midrule
         SenseAct-NN \cite{khanna2024goatbench} & \xmark & &6.62 & 3.11  &-    && 13.09 & 6.45 & - &&& 0.22 & 0.05 & -     \\
         PSL \cite{PSL2024} & \xmark &  &8.78 &3.30 &-      && 8.91 & 2.83 & - &&& 4.58 & 1.39 & -           \\
         OVON \cite{OVOV2024} &  \xmark &  &8.18 & 5.24 &-      && \textbf{15.88} & \textbf{11.35} & - &&&2.61 & 1.29 & -         \\
         VLFM \cite{VLFM2024} & \xmark&  &0.36& 0.28  &-      && 0.00 & 0.00 & - &&&0.00 & 0.00 & -         \\ \midrule
         {AIUTA} \cite{taioli2025coin} &  \mycheckmark &  & 10.46 & 4.96 & 1.31 && 14.89 & 9.11 & \textbf{1.48}  &&& 7.83 & 4.38 & 1.37     \\
          \colorbox{grayish}{\ourmethod} & \mycheckmark  & &  \textbf{12.20} & \textbf{6.32} & \textbf{1.29}   & & 14.15 & 8.29 & 1.65   &&& \textbf{9.63} & \textbf{4.55} & \textbf{1.35} &     \\
          \bottomrule
\end{tabular}}
\label{table:main_results}
\end{table*}

We adopt here the evaluation protocol for navigation discussed in \cref{sec:navbench}.

\noindent\textbf{Baseline.}
We compare \ourmethod against state-of-the-art ION baselines, including end-to-end trained policies such as SenseAct-NN \cite{khanna2024goatbench} and OVON \cite{OVOV2024}, as well as modular zero-shot approaches like PSL \cite{PSL2024} and VLFM \cite{VLFM2024}.
For CoIN, we compare against AIUTA~\cite{taioli2025coin}, the current state-of-the-art modular zero-shot method and the most direct competitor to our approach.
Extensive descriptions of these baselines in the \suppmat (\cref{sec:appendix_results}).

\noindent
\textbf{Metrics and Implementation Details.} 
We evaluate using Success Rate (\sr$\uparrow$), Success weighted by Path Length (\spl$\uparrow$), and average number of questions (\nq$\downarrow$).
Following~\cite{taioli2025coin}, an episode is considered successful if the agent reaches within $\tau = 0.25\text{m}$ of the target within $T = 500$ steps. We use the continuous environment provided by Habitat 3.0~\cite{ICCV_habitat}, based on HM3D scans~\cite{hm3d_sem}. During simulation, the user is simulated by a VLM, \texttt{Qwen3-VL-30B-A3B}~\cite{Qwen3VL}, which has exclusive access to the target image. Both the target image and the agent’s observations $O_t$ are represented at a resolution of $512 \times 512$.

\noindent
\textbf{Results.} 
\cref{table:main_results} shows the performance of all the methods on the instance object navigation episodes (see \cref{sec:navbench}), separated as in the original Goat-bench split, as \textit{Val Seen}, \textit{Val Seen Synonyms}, and \textit{Val Unseen}. 
\ourmethod significantly outperforms all methods, especially in the \textit{Val Seen} and \textit{Val Unseen} splits, while using an interaction model that 70$\times$ faster to generate questions and uses 3$\times$ less parameters as showed in \cref{tab:annotations} (Time and Size columns).
Specifically, the improvement on the \textit{Val Unseen} split is important because it shows better adaptability to unseen scenes and objects than other methods, which is a key capability for CoIN agents.
In terms of the number of questions (\nq), our method asks almost the same number of questions as AIUTA. \cref{tab:annotations} also shows the impact of the annotation quality improvement on the \sr performance of both AIUTA and \ourmethod. Both models improve by around 1.30x. %

\begin{table*}[]
\caption{Impact of annotations quality, inference time, and parameter size. The new annotations improve the results across all splits. We report the \texttt{SR} $\uparrow$ across all splits, the average inference time during interaction (\texttt{Time} $\downarrow$), and model size (\texttt{Size} $\downarrow$).}
\centering
\resizebox{0.95\textwidth}{!}{
\begin{tabular}{lc|ccc|cc}
\toprule

\textbf{Method} & \textbf{Ann. Type} 
& \textit{Val Seen} 
& \textit{Val Seen Synonyms} 
& \textit{Val Unseen} 
& \textbf{Time} $\downarrow$ & \textbf{Size} $\downarrow$\\

\midrule

\multirow{2}{*}{AIUTA \cite{taioli2025coin}}
& CoIN-Bench & 7.42  & 14.38  & 6.67 & \multirow{2}{*}{185.79} & \multirow{2}{*}{20B}  \\
& Ours & \textbf{10.46} {\scriptsize {(1.41x)}}  & \textbf{14.89} {\scriptsize {(1.04x)}}  & \textbf{7.83} {\scriptsize (1.17x)} & & \\
\midrule

\multirow{2}{*}{\colorbox{grayish}{\ourmethod}}
& CoIN-Bench & 9.14  & 11.64  & 6.43 & \multirow{2}{*}{\textbf{2.49}} & \multirow{2}{*}{\textbf{7B}}\\
& Ours & \textbf{12.20} {\scriptsize {(1.33x)}}  & \textbf{14.15} {\scriptsize {(1.33x)}} & \textbf{9.63} {\scriptsize {(1.50x)}} & &  \\

\bottomrule
\end{tabular}}

\label{tab:annotations}
\end{table*}

\vspace{-2em}

\section{Conclusions}
We introduced \ourqabenchmark, the first benchmark for CoIN that proposes separate, standardized protocols to evaluates the agent's capability in navigation and question asking. We created \ourdataset, a large-scale collection of quality-checked reasoning and question-asking traces to support training, evaluation and analysis. The dataset will be made publicly available. Using this setting, we answered one concrete modeling design question, namely that strong collaborative behavior can be achieved without heavy modular pipelines, and that a lightweight interaction-reasoning model (like \ourmethod) can be competitive while being substantially more efficient. 
More broadly, \ourqabenchmark enables further model-design questions to be systematically studied: (i) How much explicit supervision from reasoning traces is actually needed to learn effective question asking? (ii) Does coupling question generation with calibrated uncertainty estimates improve ask-or-act decisions and robustness under noisy observations or missing detections? %

\section*{Acknowledgements}
We gratefully acknowledge Gianni Franchi for his valuable input and insightful discussions during the initial phases of the project. We are also extremely grateful to Google and LambdaLabs for providing computing credits through the Google Cloud Research Credits Program and the Lambda Research Grant Application. We also acknowledge ISCRA for awarding this project access to the LEONARDO supercomputer, owned by the EuroHPC Joint Undertaking, hosted by CINECA (Italy). This work was also supported by the Ministero delle Imprese e del Made in Italy (IPCEI Cloud DM 27 giugno 2022 – IPCEI-CL-0000007) and European Union (Next Generation EU). Lastly, we acknowledge financial support from PNRR MUR project PE0000013-FAIR and PR Veneto FESR 2021-2027 24729-002238 - SINERGHY.

\bibliographystyle{splncs04}
\bibliography{main}

\clearpage\section*{Supplementary Material}
\setcounter{section}{0}

This supplementary material provides additional analyses to our proposed benchmark, offering extended empirical evaluations, data quality validations, and detailed implementation information.

We begin by expanding on the experimental details and analyses presented in the main manuscript (\cref{sec:appendix_results}). 
This includes a detailed breakdown of the navigation baselines (\cref{ssec:navigation_methods}) and an extensive evaluation of question-asking protocols, specifically focusing on the performance gap between human oracles and VLMs (\cref{ssec:qa-results}). 
Furthermore, we provide additional ablation studies to justify the necessity of fine-tuning (\cref{sec:appendix_method}).

To validate the integrity of our benchmark, we provide a thorough assessment of our data generation and human-centric evaluations. 
We detail the question-asking protocol and visualize the synthesized imagery to demonstrate its high fidelity (\cref{sec:appendix_qask}). 
This is complemented by a user study centered on navigation episodes, which evaluates the quality and linguistic characteristics of our proposed descriptions (\cref{sec:appendix_coin}).

Moreover, to ensure full transparency and reproducibility, we describe the dataset collection process in more detail (\cref{sec:appendix_rquion}), and report the complete set of prompts used across all experiments (\cref{sec:appendix_prompts}).

\section{Experimental Results}\label{sec:appendix_results}

\subsection{Details of the Navigation Methods} \label{ssec:navigation_methods}

In our navigation experiments, we experimented with the following state-of-the-art methods: {SenseAct-NN}~\cite{khanna2024goatbench}, {OVON}~\cite{OVOV2024}, {PSL}~\cite{PSL2024}, {VLFM}~\cite{VLFM2024}, and {AIUTA}~\cite{taioli2025coin}. The details of each of these methods are described below.

\noindent \textbf{SenseAct-NN  \cite{khanna2024goatbench}} introduces a recurrent policy trained end-to-end with Reinforcement Learning, integrating visual observations and textual descriptions via encoders from a pre-trained vision-language model (\ie, CLIP~\cite{radford2021clip}), alongside GPS and compass data.

\noindent \textbf{OVON \cite{OVOV2024}} is a transformer-based object-navigation policy that builds a large multidimensional state representation at each timestep by combining the current visual input, the language goal, and the previous action through two SigLIP encoders~\cite{siglip} and an action embedding layer.

\noindent \textbf{PSL \cite{PSL2024}} proposes a zero-shot policy that is pre-trained on the different tasks of image navigation, then applied to object navigation, which also uses text, using a CLIP encoder and a particular semantic perception module.

\noindent \textbf{VLFM \cite{VLFM2024}} is a fully modular zero-shot policy that builds a semantic \textit{frontier map}, assigning to each possible frontier a value indicating the likelihood of finding the object in that area using BLIP-2 \cite{blip_2}. For object detection, it uses Grounding-DINO \cite{grounding_dino} and Mobile SAM \cite{mobile_sam}. VLFM is the navigation module that we exploited in \ourmethod given its practical simplicity and good navigation performance.

\noindent \textbf{AIUTA \cite{taioli2025coin}} is the COIN method that is introduced together with COIN-Bench. AIUTA follows a modular zero-shot design, built on top of VLFM, using two large models (a VLM and an LLM) that dialogue with each other every time a potential target is detected to reduce uncertainty on visual understanding,  and understand how to interact with the user under the collaborative setting.
AIUTA is most related to \ourmethod.
\ourmethod presents a smaller, lighter single-module question-asking VLM, that outperforms the larger, heavier multi-module AIUTA.

\begin{table}
\caption{Comparison between VLM-based and human as oracle.
The questioner is \texttt{Gemini-3-flash} \cite{deepmind_gemini3flash_2025}. In the first row, we use three different VLMs as oracles following the same evaluation in the main manuscript. In the second row, we task humans to answer the questions in real-time while using \ourmethod. }
\centering
\scriptsize
\begin{tabular}{ccccccccccccc}
\toprule %
\multicolumn{1}{c}{Oracle} 
& \multicolumn{2}{c}{\textit{Cat}} 
& \multicolumn{2}{c}{\textit{CatCol}} 
& \multicolumn{2}{c}{\textit{ColFeat}} 
& \multicolumn{2}{c}{\textit{Ctx}} 
& \multicolumn{2}{c}{\textit{ColCtx}} 
& \multicolumn{2}{c}{ColCtxFeat} \\
\cmidrule(lr){1-1}
\cmidrule(lr){2-3}
\cmidrule(lr){4-5}
\cmidrule(lr){6-7}
\cmidrule(lr){8-9}
\cmidrule(lr){10-11}
\cmidrule(lr){12-13}
Human 
& \textbf{SR}$\uparrow$ & \textbf{FR}$\uparrow$
& \textbf{SR}$\uparrow$ & \textbf{FR}$\uparrow$
& \textbf{SR}$\uparrow$ & \textbf{FR}$\uparrow$
& \textbf{SR}$\uparrow$ & \textbf{FR}$\uparrow$
& \textbf{SR}$\uparrow$ & \textbf{FR}$\uparrow$
& \textbf{SR}$\uparrow$ & \textbf{FR}$\uparrow$ \\
\midrule
\xmark 
& 0.93  &  0.90
& 0.92  &  0.88
& 0.93  &  0.90
& 0.90  &  0.83
& 0.95  &  0.93
& 0.96  &  0.93 \\ 
\mycheckmark 
& 0.92  &  0.88
& 0.90  &  0.85
& 0.92  &  0.87
& 0.90  &  0.86
& 0.96  &  0.94
& 0.94  &  0.87 \\
\bottomrule
\vspace{1em}
    \end{tabular}
    \label{tab:results_qask_human}
\end{table}

\begin{table*}[t]
\caption{Success Rate (\texttt{SR} $\uparrow$) and Finish Rate (\texttt{FR} $\uparrow$) across different annotation types of \ourqabenchmark. See main manuscript, \cref{subsec:q-a} for details about the annotation types.}
\centering
\scriptsize
\begin{tabular}{lcc|cc|cc|rc|cc|cc}
\toprule
& \multicolumn{2}{c|}{Cat} 
& \multicolumn{2}{c|}{CatCol} 
& \multicolumn{2}{c|}{ColFeat} 
& \multicolumn{2}{c|}{Ctx} 
& \multicolumn{2}{c|}{ColCtx} 
& \multicolumn{2}{c}{ColCtxFeat} \\
\cmidrule(lr){2-3}
\cmidrule(lr){4-5}
\cmidrule(lr){6-7}
\cmidrule(lr){8-9}
\cmidrule(lr){10-11}
\cmidrule(lr){12-13}
\textbf{Method} 
& \textbf{SR}$\uparrow$ & \textbf{FR}$\uparrow$
& \textbf{SR}$\uparrow$ & \textbf{FR}$\uparrow$
& \textbf{SR}$\uparrow$ & \textbf{FR}$\uparrow$
& \textbf{SR}$\uparrow$ & \textbf{FR}$\uparrow$
& \textbf{SR}$\uparrow$ & \textbf{FR}$\uparrow$
& \textbf{SR}$\uparrow$ & \textbf{FR}$\uparrow$ \\
\midrule
Qwen2.5-VL-7B\cite{Qwen25VL}
& 0.02  &  0.0
& 0.04  &  0.0
& 0.13  &  0.01
& 0.05  &  0.01
& 0.08  &  0.0
& 0.13  &  0.01\\
AIUTA$^\circ$ \cite{taioli2025coin}
& \textbf{0.42} & \textbf{0.28} 
& 0.34 & 0.22 
& 0.30 & 0.19 
& 0.32 & 0.22 
& 0.24 & 0.15 
& 0.20 & 0.13 \\

\colorbox{grayish}{\ourmethodc} 
&0.06 & 0.03
& \textbf{0.45} & \textbf{0.31}
& \textbf{0.58} & \textbf{0.44}
& \textbf{0.43} & \textbf{0.32}
& \textbf{0.61} & \textbf{0.42}
& \textbf{0.68} & \textbf{0.54}\\
\midrule
Qwen3-VL-30B-A3B \cite{Qwen3VL}
&0.02 & 0.01
& 0.24 & 0.18
& 0.40 & 0.34
& 0.34 & 0.29
& 0.47 & 0.41
& 0.54 & 0.47\\
Gemini-3-flash \cite{deepmind_gemini3flash_2025}
& \textbf{0.93}  &  \textbf{0.90}
& \textbf{0.92}  &  \textbf{0.88}
& \textbf{0.93}  &  \textbf{0.90}
& \textbf{0.90}  &  \textbf{0.83}
& \textbf{0.95}  &  \textbf{0.93}
& \textbf{0.96}  &  \textbf{0.93}\\
\bottomrule
\end{tabular}

\label{tab:qask_results_complete}
\end{table*}

\subsection{Human Evaluation and Results on Question-Asking} \label{ssec:qa-results}

\begin{figure}
    \centering
    \includegraphics[width=0.58\linewidth]{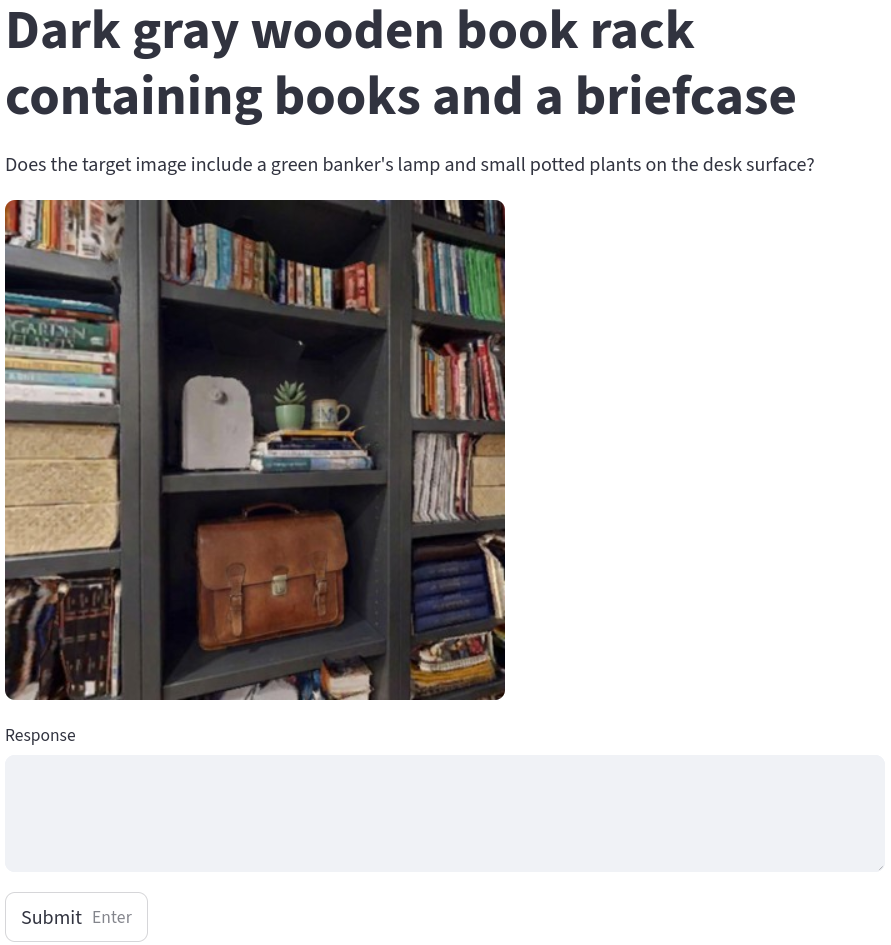}
    \caption{Tool for the human oracle tests. The target image is presented to the user, along with the task description (the title), and the current question asked by the questioner (below the title). The user is instructed to answer by writing the answer in the text box below. }
    \label{fig:human_oracle_tool}
\end{figure}

\noindent \textbf{VLM-based \textit{v.s.} Human Evaluation.} 
Our evaluation protocol includes VLM-based oracles to make experimentation fast, scalable and reproducible. 
However, a fair question raises: \textit{How does the model behave when a real human takes the role of the oracle?}
In addition, we want to investigate whether the proposed scalable way to run the evaluation without costly human intervention is a viable and consistent way to test for question-asking capabilities, without introducing errors and biases in the tests. 
To answer these questions, we tested the state-of-the-art model \texttt{Gemini-3-flash} as a questioner, and have human annotators as oracles.
\cref{tab:results_qask_human} shows the results of the VLM-based oracle vs. the human oracle. 
Under different target description types, we observe consistently similar results in terms of SR and FR obtained by both oracle types. 
Moreover, we find that, on average, a human response can take from 4 seconds up to 15, whereas a VLM takes as little as 2 seconds, with less variability. 
Moreover, using an VLM oracle allows us to run multiple evaluations in parallel, which further scales up the evaluation, that is not feasible with human evaluation.
This demonstrates that our VLM-based oracle can serve as a viable alternative to simulate costly human responses, significantly scaling up experimentation and speeding up the process tenfold.
\cref{fig:human_oracle_tool} shows the tool used to collect the human answers.

\noindent \textbf{Additional Question-asking Results.} 
\cref{tab:qask_results2} of the main paper shows the performance of question-asking methods on different annotation types.
We extend that table by adding the pre-trained base model (\texttt{Qwen2.5-VL-7B}) and the larger non-navigation-ready methods (\texttt{Qwen3-VL-30B-A3B}) in \cref{tab:qask_results_complete}.
We report the \texttt{SR} and \texttt{FR} across all types of descriptions. 
We can notice that the base model \texttt{Qwen2.5-VL-7B} \cite{Qwen25VL} achieves lower scores than \ourmethod$^\circ$ for all types, especially those with less information (\eg, category and context). 
Like other methods (except AIUTA$^\circ$ \cite{taioli2025coin}), \texttt{Qwen2.5-VL-7B} performance improves when it receives more information about the task (scanning the corresponding row of the table from left to right).
On the other hand, \texttt{Gemini-3-flash} \cite{deepmind_gemini3flash_2025} exhibits superior question-asking capabilities across all types, reaching 96\% on the most informative type (color-context-feature, last column). 
It is worth noting that this evaluation functions as an upper bound for CoIN, since \texttt{Gemini-3-flash} is estimated to be very large (order of 1 trillion parameters), thus making the model non-navigation-ready.

\section{Navigation Protocol}\label{sec:appendix_coin}

\begin{table}[t]
    \caption{Comparison between CoIN-Bench \cite{taioli2025coin} and our task descriptions evaluated by LLM-as-judge (\texttt{Gemini-3-pro}) and humans. Our task descriptions are highly preferred and exhibit a very low rate of problematic description concerning hallucination or imprecise details (2\%). In contrast, the CoIN-Bench descriptions are rarely preferred with a high problem rate (over 50\%).}
\centering
\small
\setlength{\tabcolsep}{4pt}
    \begin{tabular}{llcc}
    \toprule
    \textbf{Evaluator} & \textbf{Task Descriptions} & \textbf{Preference Rate} $\uparrow$ & \textbf{Problem Rate} $\downarrow$ \\
    \midrule
    \multirow{2}{*}{LLM as judge} & CoIN-Bench & 0.171 & 0.618 \\
    & Ours & \textbf{0.829} & \textbf{0.020} \\
    \midrule
    \multirow{2}{*}{Humans} & CoIN-Bench & 0.183 & 0.530 \\
    & Ours & \textbf{0.816} & \textbf{0.056} \\
    \bottomrule
    \end{tabular}

    \label{tab:preference_annotations}
\end{table}

\subsection{Human Evaluation of Task Descriptions}\label{sec:appendix_annotations}
In our navigation protocol, we proposed a recipe to generate high-quality and diverse task descriptions for navigation. 
Therefore, we assess the quality of the proposed task descriptions when comparing them to CoIN-Bench's ones~\cite{taioli2025coin}.
We set up a preference test with two choices (CoIN-Bench \cite{taioli2025coin} \textit{v.s.} ours, order randomized) on 50\% of the data samples and ask a evaluator which option they prefer. For this task, we use both an LLM-as-judge evaluator (\texttt{Gemini-3-pro}) and 5 human evaluators.

We instruct the evaluator to focus on \textit{precision}, \textit{conciseness}, and \textit{clarity}; moreover, we ask the evaluator to check the annotations for \textit{critical problems} concerning imprecise details and hallucinations via additional check boxes.
The evaluation tool we developed is shown in \cref{{fig:preference_annotation_tool}}.
We measure the \texttt{preference rate} ($\uparrow$) defined as the ratio between the number of preferences expressed for one or the other description and the total number of expressed preferences; and the \texttt{problem rate} ($\downarrow$) defined as the ratio between the number of problematic descriptions with hallucinations or imprecise details and the number of annotations of that type.

\begin{figure}
    \centering
    \includegraphics[width=0.58\linewidth]{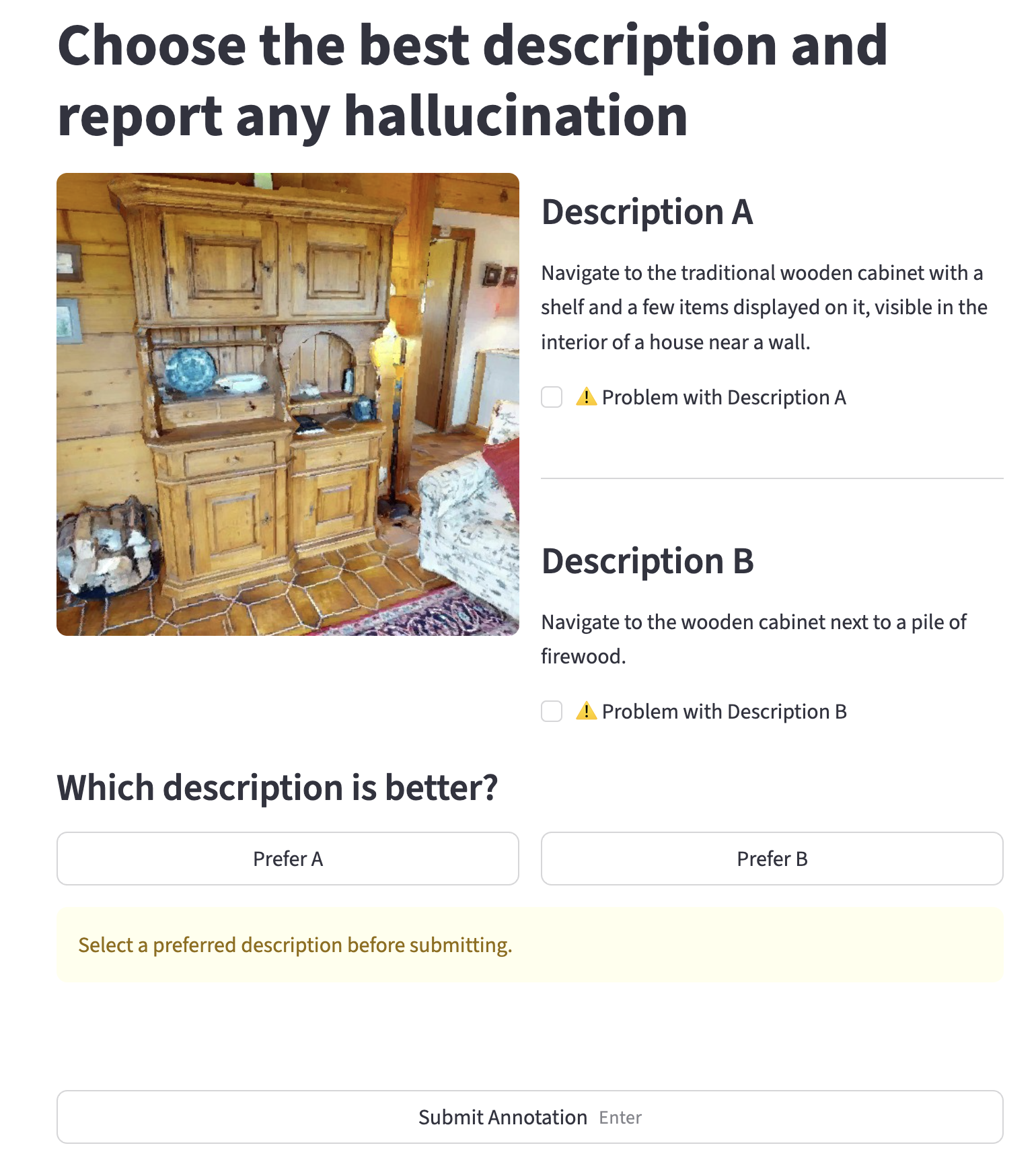}
    \caption{Evaluation tool for preference testing in our human evaluation. Two choices are presented to the user. The users are instructed to click on the checkbox ``Problem with Description X'' if they find hallucinations and clear mistakes.}
    \label{fig:preference_annotation_tool}
\end{figure}

We report the results of this study in \cref{tab:preference_annotations}, where the results of the LLM-as-judge evaluator are in the first two rows, and the results of 5 human evaluators are in the last two rows.
In both cases, we notice that our task descriptions are preferred over 80\% of the time, while CoIN-Bench descriptions exhibit a high rate of problematic descriptions concerning imprecise details or hallucinations.
Instead, the rate of problematic descriptions in our dataset is significantly lower, reducing from a very high value of 50-60\% of COIN-Bench to only 2-5\%.
We can also observe the alignment and consistency between the human and LLM-as-judge evaluators, contributing to similar ballpark results. 

\begin{figure}
    \centering
    \includegraphics[width=0.54\linewidth]{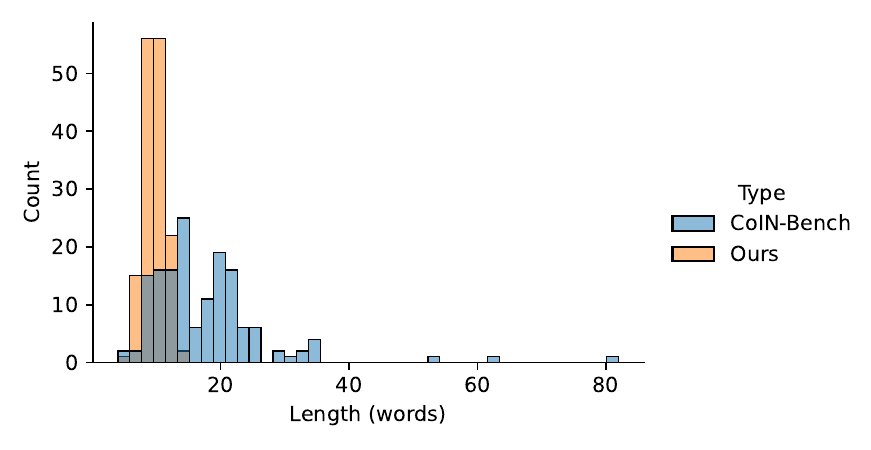}
    \caption{Length of descriptions, in words. The CoIN-Bench descriptions exhibit extreme outliers and, in general, are longer and wordier than our descriptions. }
    \label{fig:length_annotations_appendix}
\end{figure}

Finally, we analyze in \cref{fig:length_annotations_appendix} the distribution of the length of the task descriptions in terms of words.
As shown clearly, our task descriptions are shorter with many fewer outliers, compared to the ones of COIN-Bench. 
The average length of CoIN-Bench descriptions is around $\sim 17$, with a standard deviation of $\sim 9$. In contrast, our annotations are $\sim 9$ words long with a smaller standard deviation of $\sim2$.

\subsection{Episode Data}

\begin{figure}
    \centering
    \includegraphics[width=0.75\linewidth]{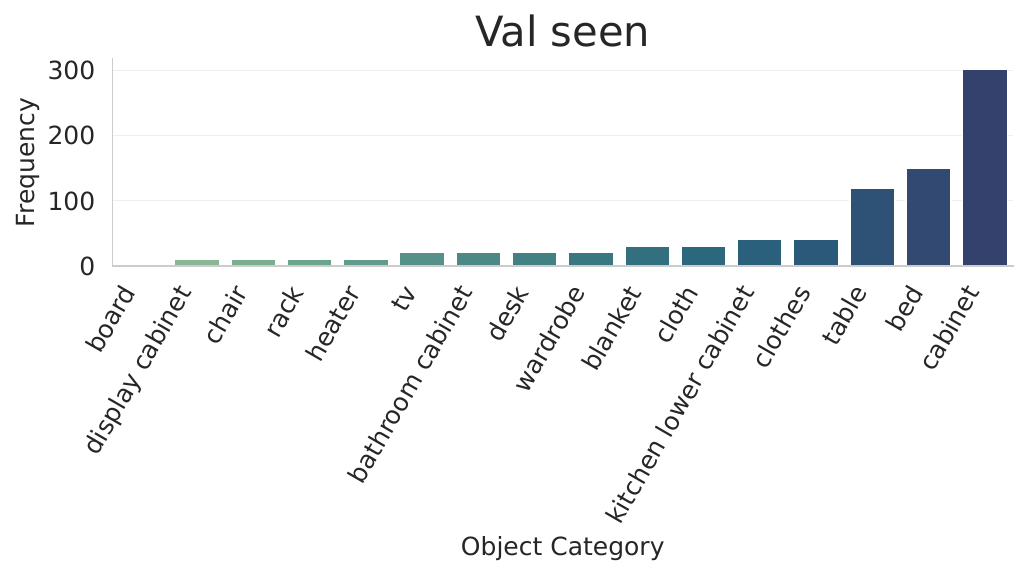}
    \includegraphics[width=0.75\linewidth]{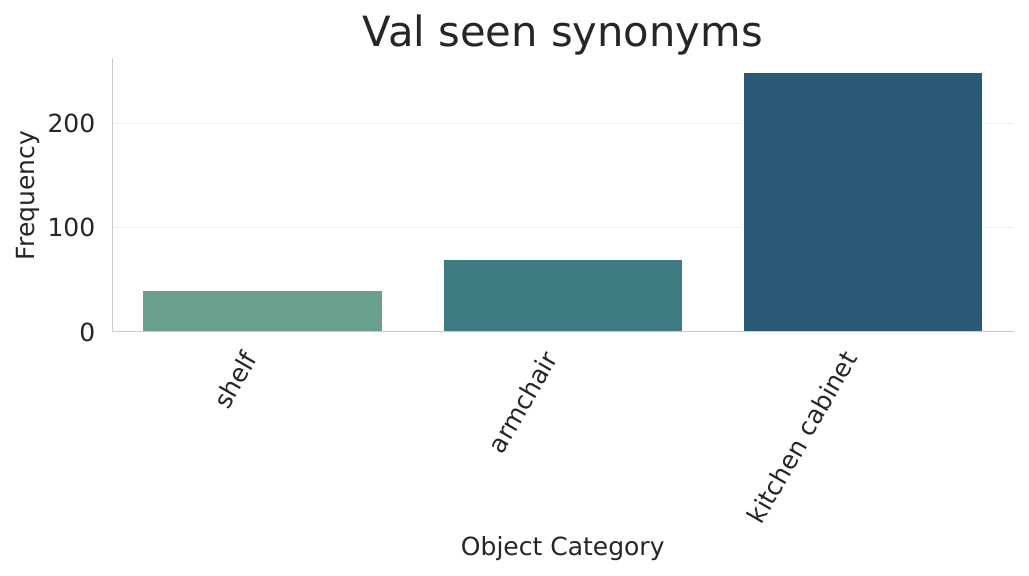}
    \includegraphics[width=0.75\linewidth]{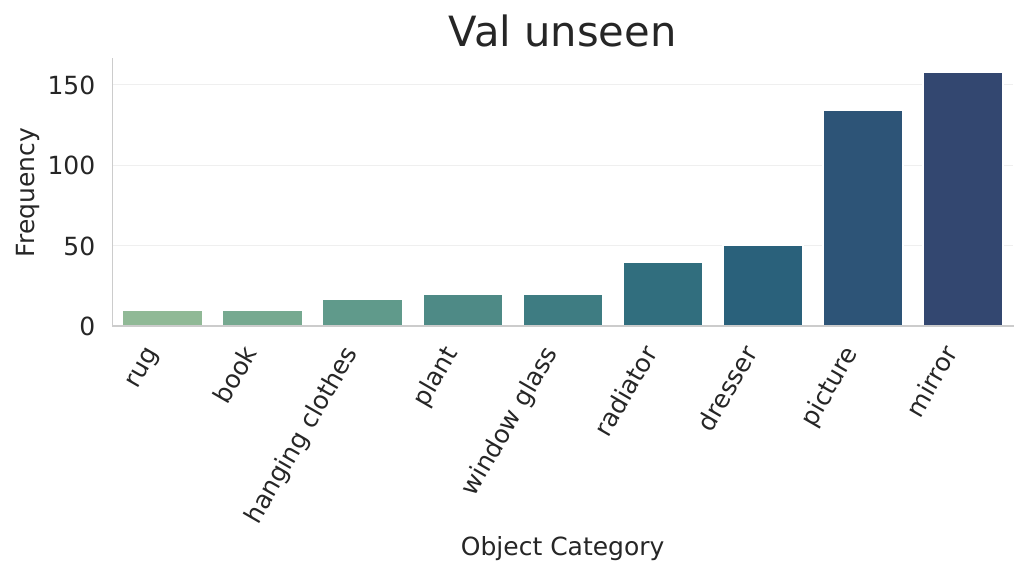}
    \caption{Distribution of the target instance objects categories, per split, for the navigation episodes.}
    \label{fig:navigation_distribution}
\end{figure}

Here we provide more information about the navigation episodes. In \cref{fig:navigation_distribution}, we report the distribution of object categories for all goal instance objects of our navigation episodes, divided into the three splits. The data split follows prior benchmarks' practice to maintain compatibility. We can see that the \textit{val seen synonyms} split only contains three categories of objects, whereas the \textit{train} and \textit{val unseen} split contain at least 9 different categories. 

\section{Question-asking Protocol}\label{sec:appendix_qask}

\subsection{Object Category Distribution in Episode Data}

\begin{figure}[t]
    \centering
    \includegraphics[width=1\linewidth]{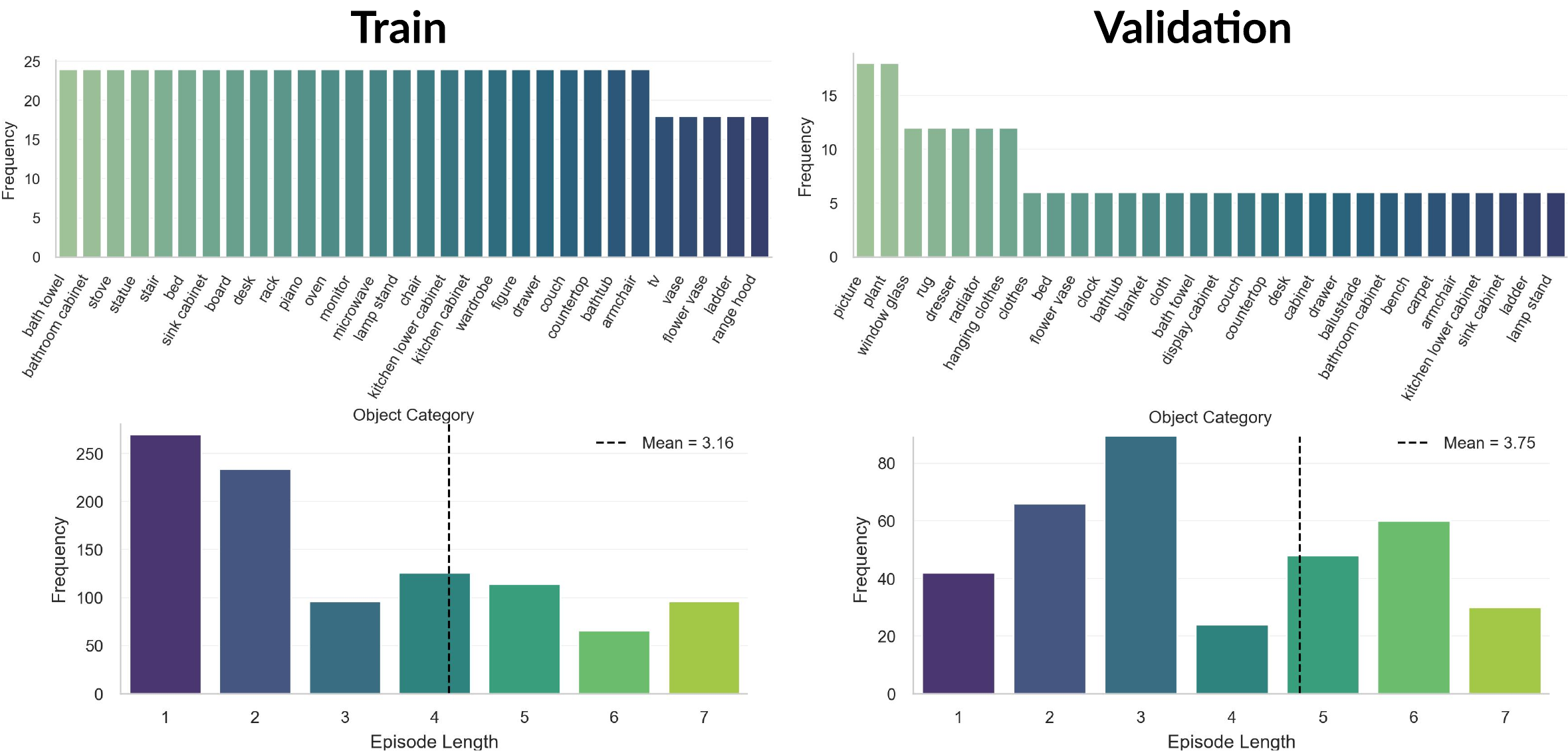}
    \caption{Categories distribution (top 30) of the target image, and lengths of episodes (in number of distractors; dotted line is the mean length), for the question-asking protocol (left: \textit{Train} split, right: \textit{Validation} split). The categories are well distributed, with the top 30 having at least 16 (for the train split) or 6 (for the validation split) examples. }
    \label{fig:coin_bench_data}
\end{figure}

\cref{fig:coin_bench_data} reports statistics of the question-asking episode data. On the left, we report the object category distribution (top) and the length of episodes, as measured in terms of the number of observations (bottom), for the training episodes. 
On the right, we report the same statistics for the validation episodes. 
For the category distribution, we report data of the top-30 most frequent categories for best visibility.
We can see that the object categories are well-distributed, especially for the training episodes, with the top-30 categories each having more than 16 episodes. In terms of episode length, the average validation episode is slightly longer than the average training episode at $\sim4.8$ instead of around $\sim4.1$.

\subsection{Augmentation of Images and Descriptions with Editing}
\begin{figure}
    \centering
    \includegraphics[width=0.24\linewidth]{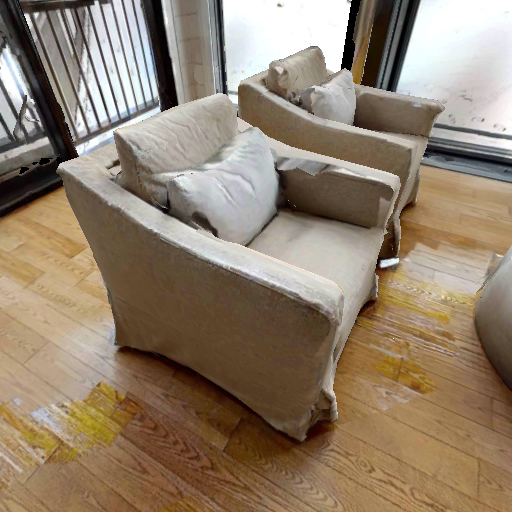}
    \includegraphics[width=0.24\linewidth]{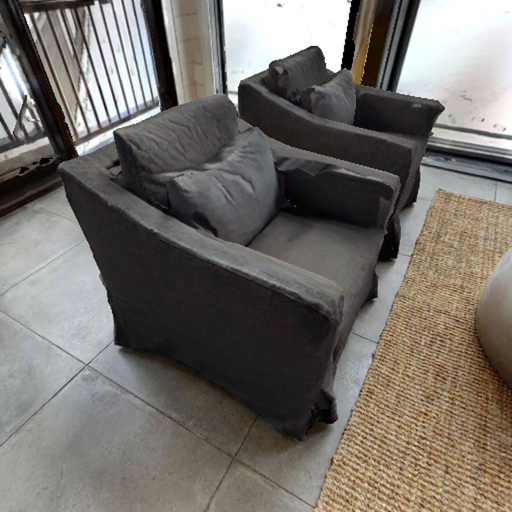}
    \includegraphics[width=0.24\linewidth]{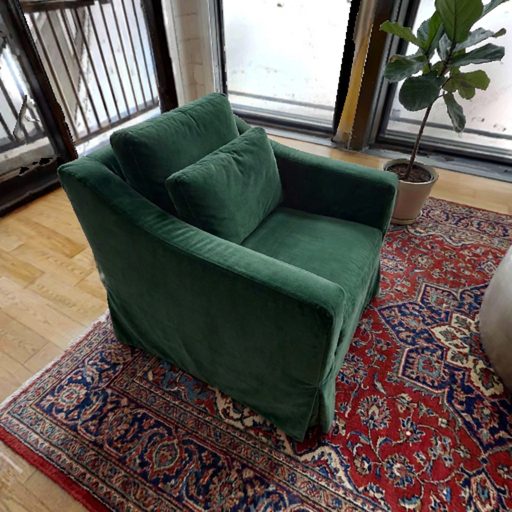}
    \includegraphics[width=0.24\linewidth]{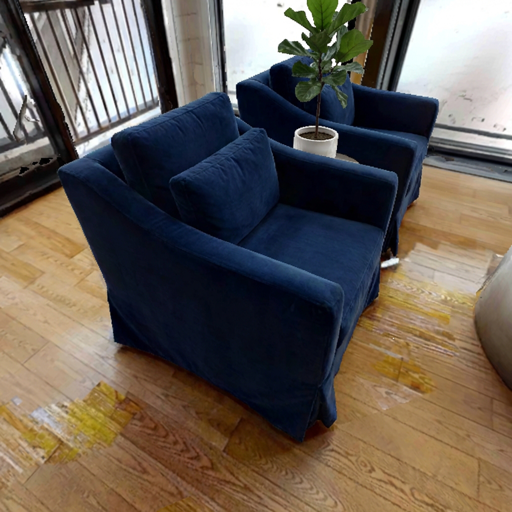}\vspace{2em}
    \includegraphics[width=0.24\linewidth]{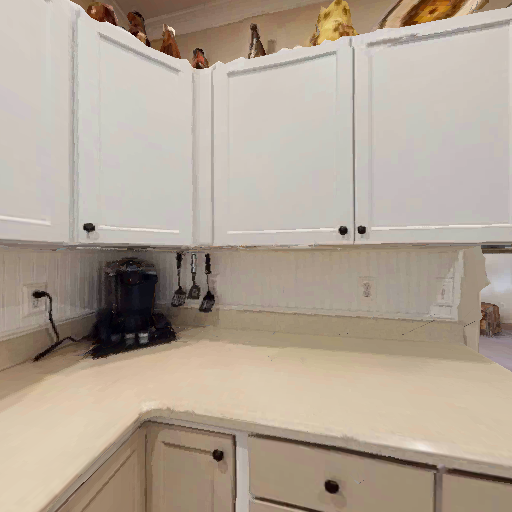}
    \includegraphics[width=0.24\linewidth]{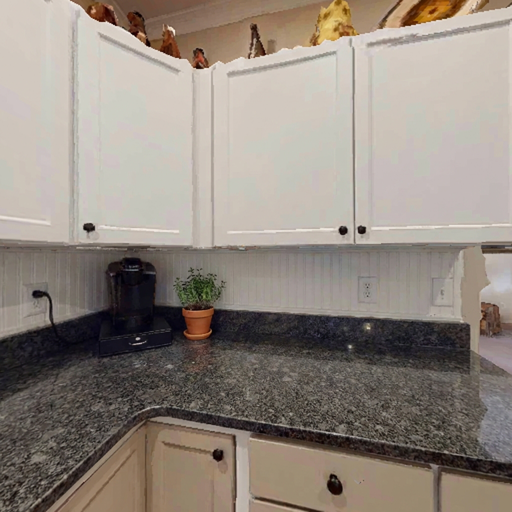}
    \includegraphics[width=0.24\linewidth]{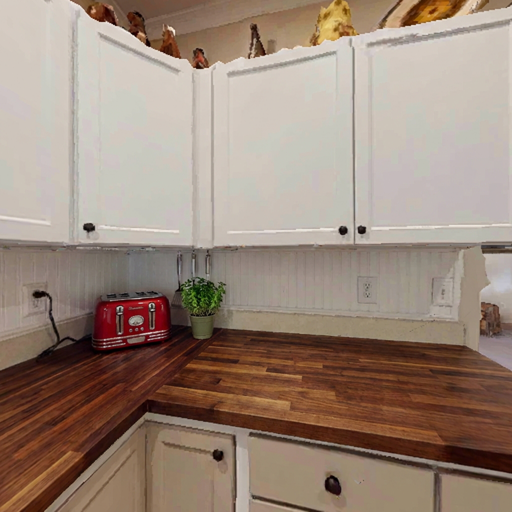}
    \includegraphics[width=0.24\linewidth]{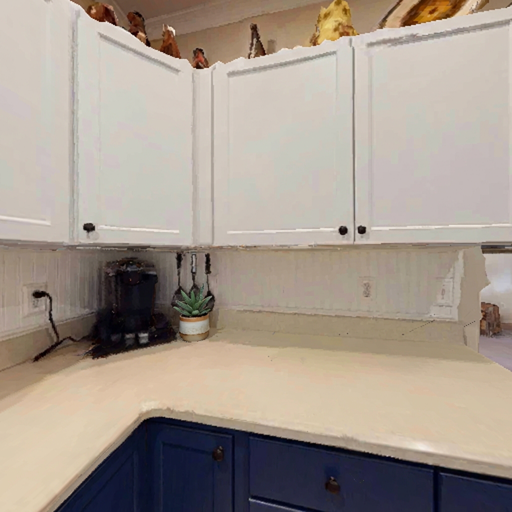}\vspace{2em}
    \includegraphics[width=0.24\linewidth]{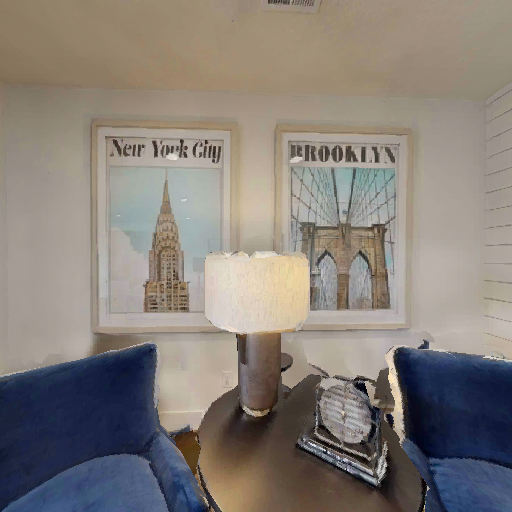}
    \includegraphics[width=0.24\linewidth]{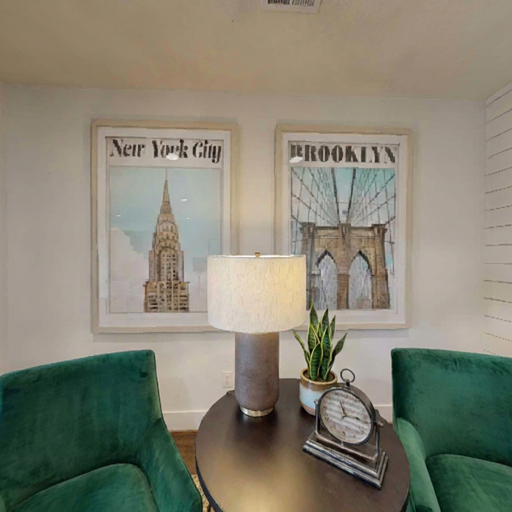}
    \includegraphics[width=0.24\linewidth]{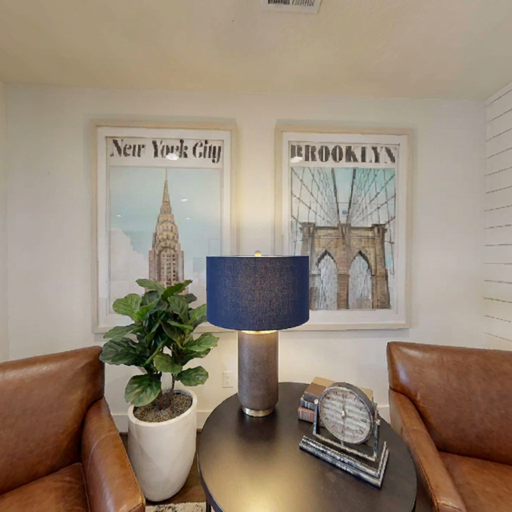}
    \includegraphics[width=0.24\linewidth]{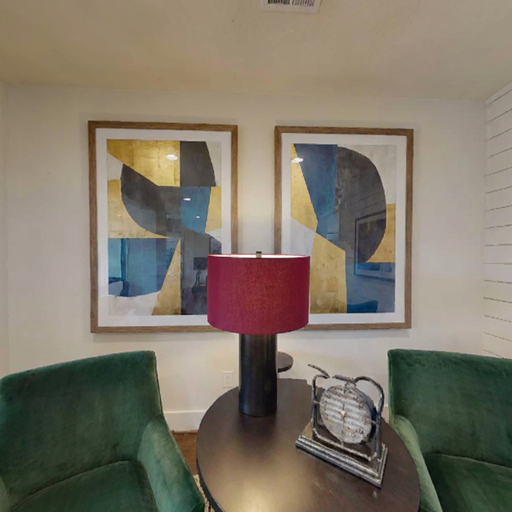}\vspace{2em}
    \includegraphics[width=0.24\linewidth]{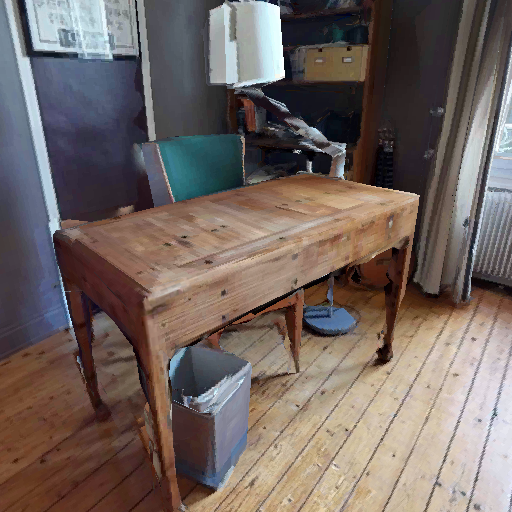}
    \includegraphics[width=0.24\linewidth]{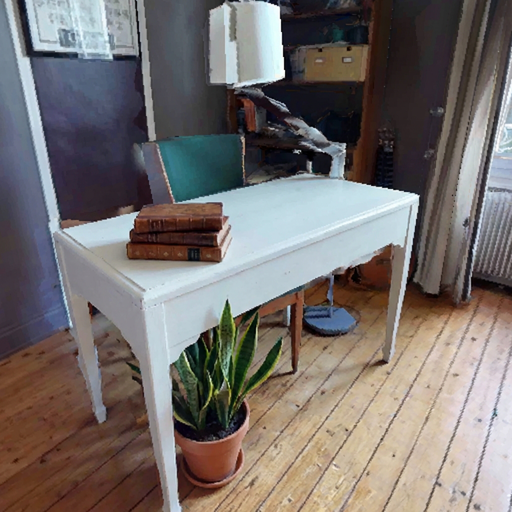}
    \includegraphics[width=0.24\linewidth]{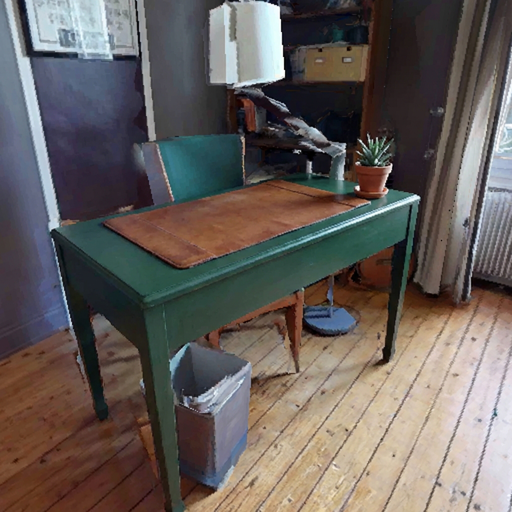}
    \includegraphics[width=0.24\linewidth]{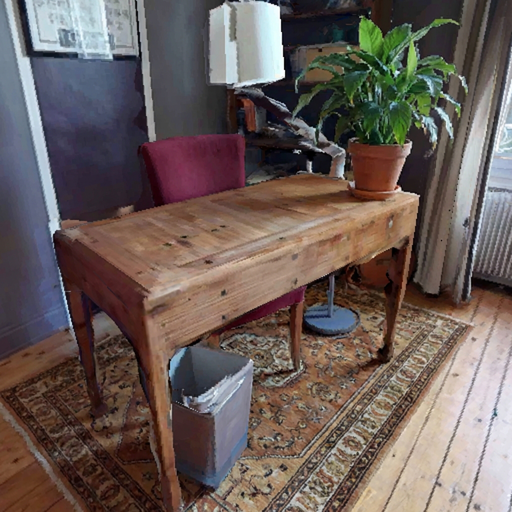}
    \caption{Variability of images used in our question-asking protocol. The first image is the original image; the next three are variations obtained using a strong generative model (\texttt{Gemini-3-pro-image} \cite{deepmind_gemini3proimage})}
    \label{fig:image_variation}
\end{figure}

\begin{figure}
    \centering
    \includegraphics[width=0.24\linewidth]{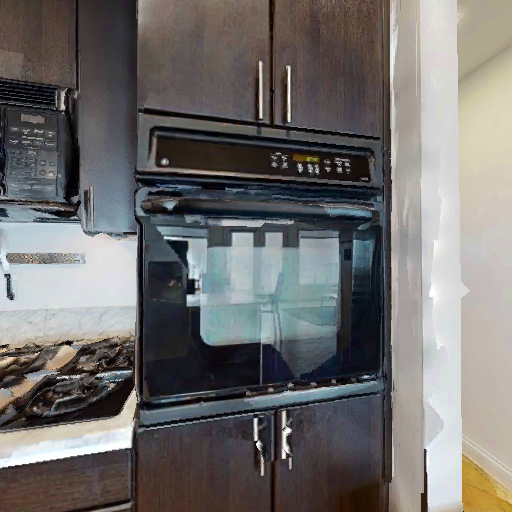}
    \includegraphics[width=0.24\linewidth]{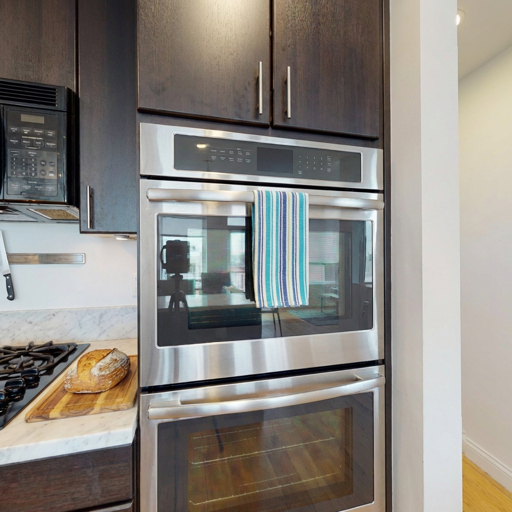}
    \includegraphics[width=0.24\linewidth]{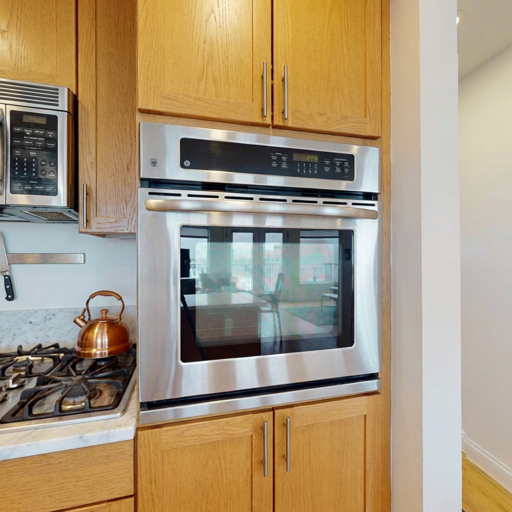}
    \includegraphics[width=0.24\linewidth]{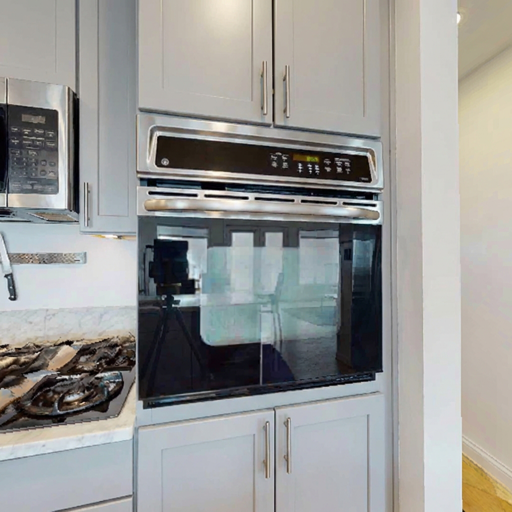}\vspace{2em}
    \includegraphics[width=0.24\linewidth]{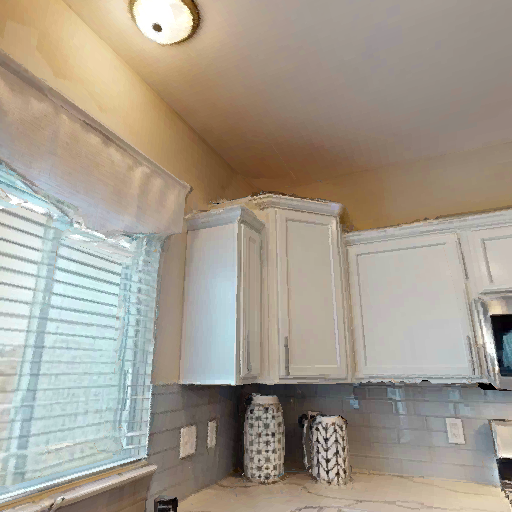}
    \includegraphics[width=0.24\linewidth]{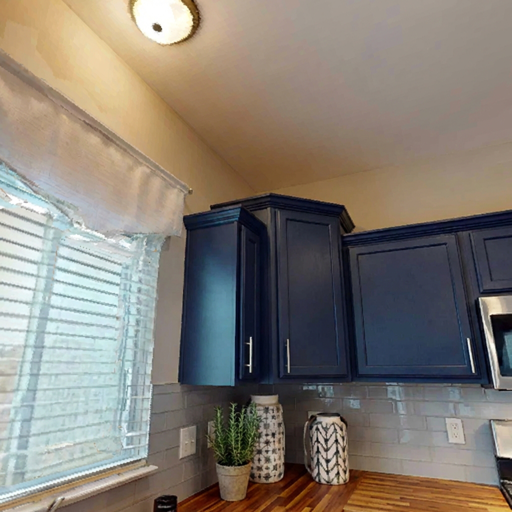}
    \includegraphics[width=0.24\linewidth]{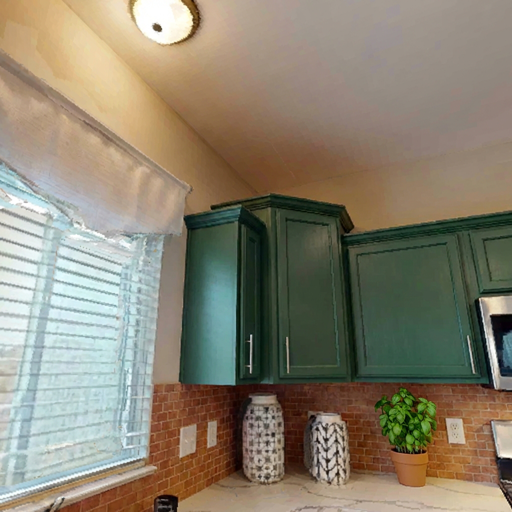}
    \includegraphics[width=0.24\linewidth]{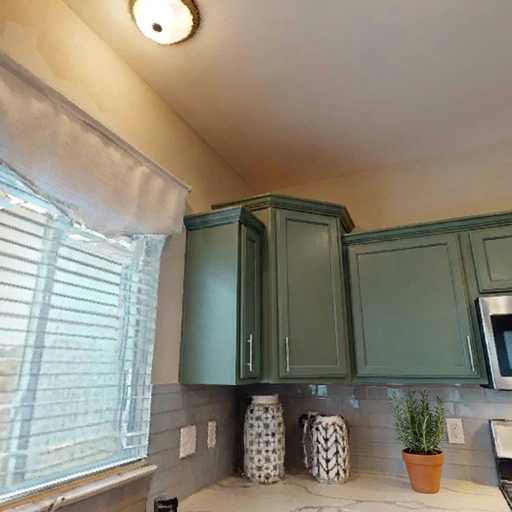}\vspace{2em}
    \includegraphics[width=0.24\linewidth]{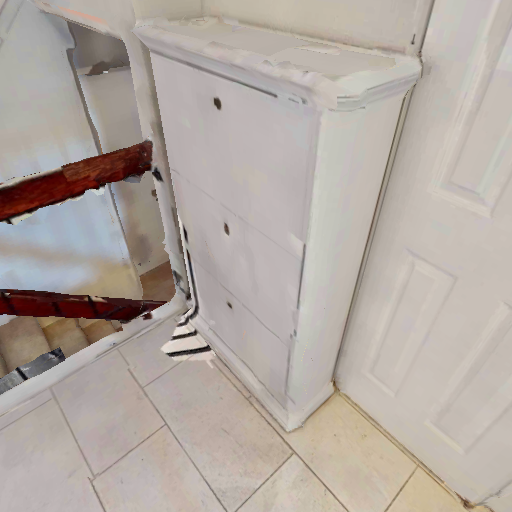}
    \includegraphics[width=0.24\linewidth]{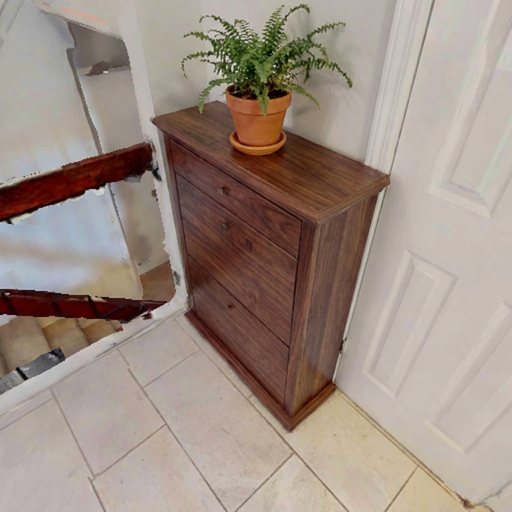}
    \includegraphics[width=0.24\linewidth]{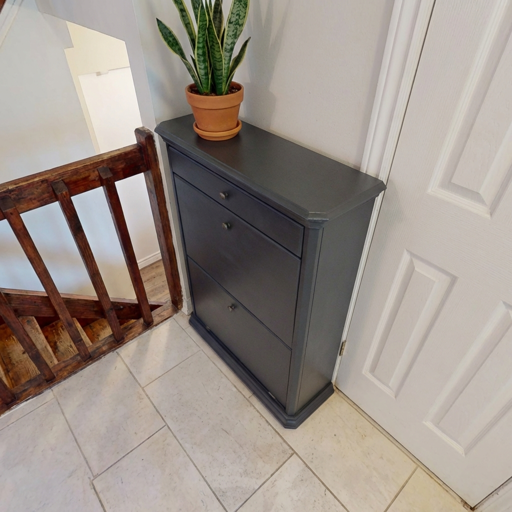}
    \includegraphics[width=0.24\linewidth]{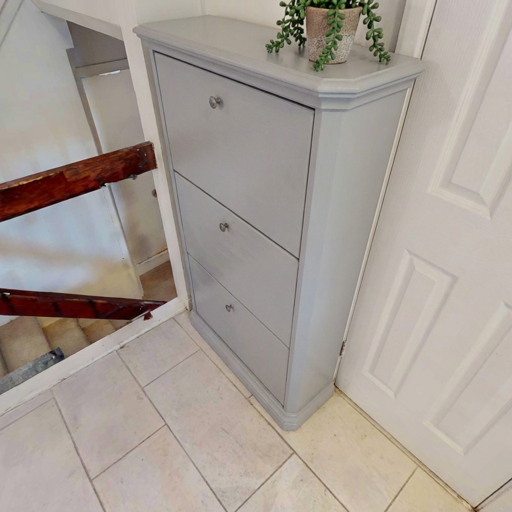}\vspace{2em}
    \includegraphics[width=0.24\linewidth]{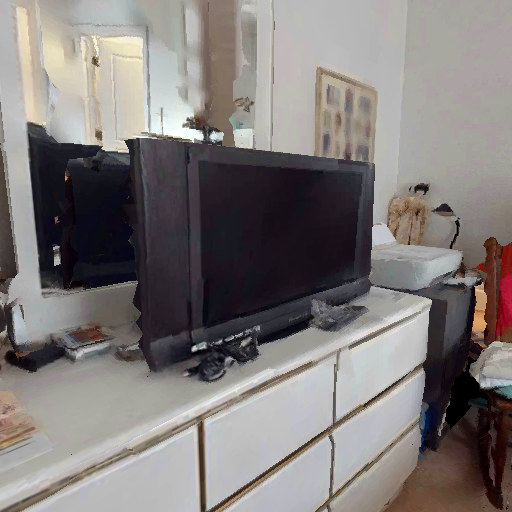}
    \includegraphics[width=0.24\linewidth]{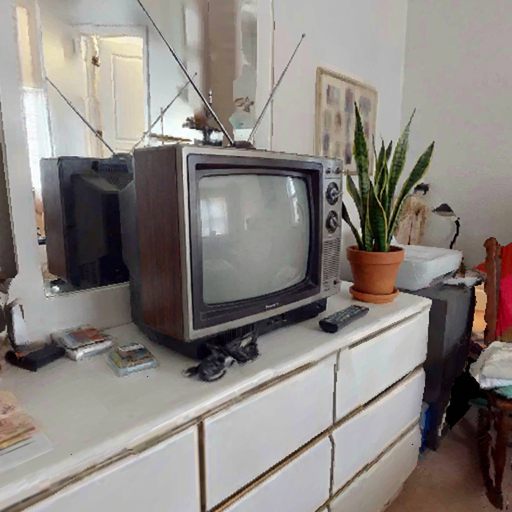}
    \includegraphics[width=0.24\linewidth]{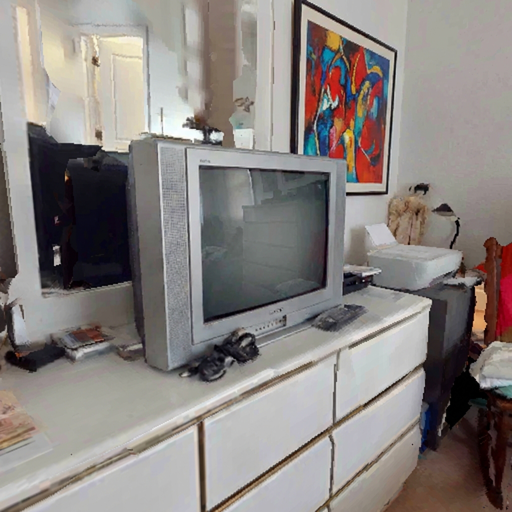}
    \includegraphics[width=0.24\linewidth]{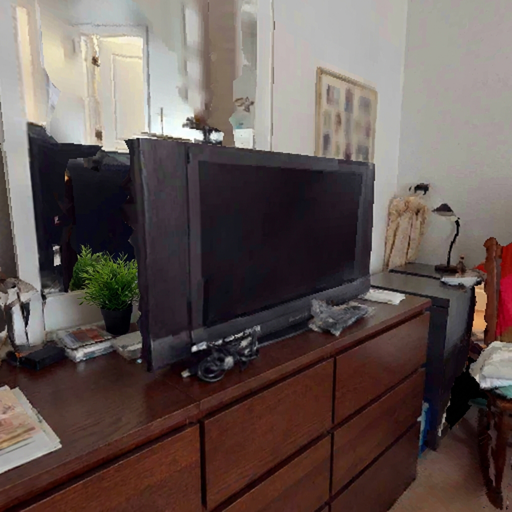}
    \caption{Variability of images used in our question-asking protocol (cont'd).}
    \label{fig:image_variation2}
\end{figure}

\begin{figure}[t]
\centering
\includegraphics[width=0.48\linewidth]{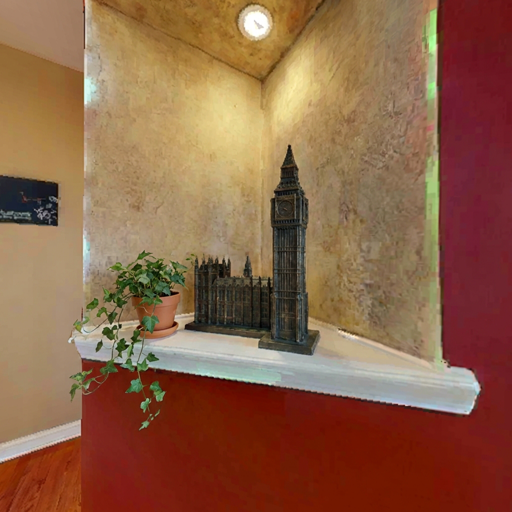}\\
\vspace{4pt}
\small
\begin{tabular}{lcl}
\textbf{Category} & &Statue \\
\textbf{Category-Color} & &Dark bronze statue \\
\textbf{Color-Feature} & &Dark bronze statue of Big Ben \\
\textbf{Context} & &Statue standing on a shelf next to a potted plant \\
\multirow{2}{*}{\textbf{Color-Context}} & &Dark bronze statue standing on a shelf \\
& & next to a potted plant \\
\multirow{2}{*}{\textbf{Color-Context-Feature}} & &Dark bronze Big Ben statue standing on a shelf\\
& & next to a potted plant \\
\vspace{1em}
\end{tabular}
\includegraphics[width=0.48\linewidth]{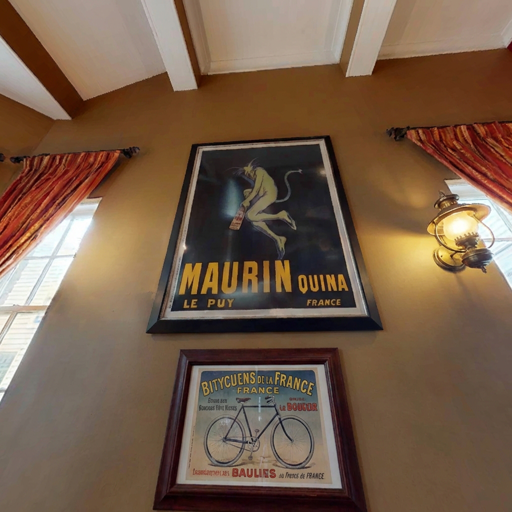}\\
\vspace{4pt}
\small
\begin{tabular}{lcl}
\textbf{Category} & &Picture \\
\textbf{Category-Color} & &Black framed poster \\
\textbf{Color-Feature} & &Black framed poster with yellow text \\
\textbf{Context} & &Poster hanging on a wall above a bicycle print \\
\multirow{2}{*}{\textbf{Color-Context}} & &Black poster hanging on a brown \\
& & wall above a print \\
\multirow{2}{*}{\textbf{Color-Context-Feature}} & &Black framed poster with yellow text \\
& & on a brown wall above a print \\
\end{tabular}
\caption{Different types of generated descriptions.}
\label{fig:different_descriptions1}
\end{figure}

\begin{figure}
    \centering
\includegraphics[width=0.48\linewidth]{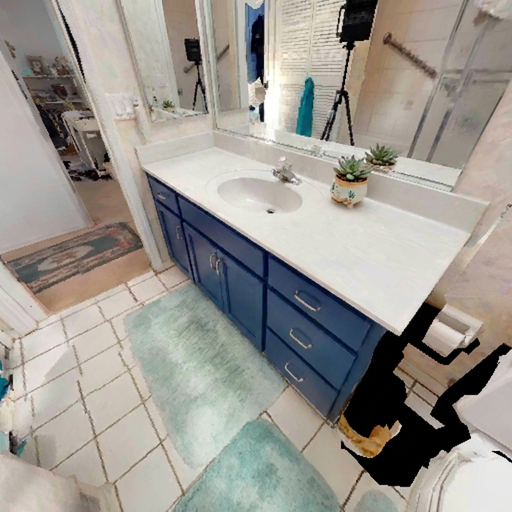}\\
\vspace{4pt}
\small
\begin{tabular}{lcl}
\textbf{Category} & &Bathroom cabinet \\
\textbf{Category-Color} & &Dark blue cabinet \\
\textbf{Color-Feature} & &Dark blue cabinet with silver handles \\
\textbf{Context} & &Cabinet underneath a white sink \\
\multirow{1}{*}{\textbf{Color-Context}} & &Dark blue cabinet underneath a white sink \\
\multirow{2}{*}{\textbf{Color-Context-Feature}} & &Dark blue cabinet with silver handles \\
& & underneath a white sink \\
\end{tabular}
\caption{Different types of generated descriptions (cont'd).}
\label{fig:different_descriptions2}
\end{figure}

Relying solely on images obtained from navigation episodes is limiting, because they can be of low quality, have bad viewpoints, or be very different from each other.
Therefore, we start from images of common household objects and use a state-of-the-art commercial image generation model (\texttt{Gemini-3-pro-image} \cite{deepmind_gemini3proimage}) to edit the original images to differ in a small number of details.
For each original image, we generate five additional images, changing 1 to 5 details. These details include color, shape, texture, presence, and position of objects. 
Moreover, we add or remove objects from the images. 
\cref{fig:image_variation} and \cref{fig:image_variation2} show 32 examples of images generated using our protocol. 
On the left, we show the original image; on the right, three synthetic images derived from it. The results are of high visual quality with coherent semantics. The changed details are mostly color and surrounding objects, but also involve texture and shape. 
In addition, we generate six different descriptions for the images, differing in terms of information density and specificity. See \cref{fig:different_descriptions1} and \cref{fig:different_descriptions2} for three examples of different description types.

\section{\ourdataset}\label{sec:appendix_rquion}

\subsection{Data Collection}

We describe here how we built \ourdataset, which was done in two independent phases to increase diversity.

In the first phase, we collected around 500 instance object images from both our navigation episodes and other related datasets  (GOAT-Bench \cite{khanna2024goatbench} and CoIN-Bench \cite{taioli2025coin}) to have a diverse set of objects and scenarios. 
We created three splits sourced from the original navigation datasets~\cite{khanna2024goatbench,taioli2025coin}: (i) \textit{Train} contains images sourced from train episodes; (ii) \textit{Val seen} has images from \textit{Val seen} and \textit{Val seen synonyms} episodes, and (iii) \textit{Val unseen} contains images sourced from validation episodes including objects not present in other splits.
Human annotators are tasked to compose two high-quality descriptions for each image.

We then group all images and descriptions by object category, creating a pool of images and associated descriptions.
In addition, we used \texttt{Gemini-3-flash} \cite{deepmind_gemini3flash_2025} to write 20 additional descriptions for each object category (not based on any image), and add these descriptions to the pool of descriptions for that object category.
Then, we prompted \texttt{Gemini-3-pro} to write high-quality reasoning $R$, scores $S$, and questions $Q$ given the description $D$ and image $O$, as described in the main manuscript in \cref{subsec:ourdataset}. 
By doing so, we obtain approximately 24,000 $(D,O,R,S,Q,C)$ samples (note: the context $C$ is empty here and used during navigation). 
In this phase, the dataset is highly imbalanced toward entries with scores 0 and 2, with very few questions. Later, in the second phase, we collect many more questions and entries with a score of $1$. 

In the second phase, we collected a set of samples following our question-asking protocol, \emph{i.e.}, we run \texttt{Gemini-3-flash} \cite{deepmind_gemini3flash_2025} as questioner using another instance of \texttt{Gemini-3-flash} as oracle. 
During each interaction, the questioner is presented with a description $D$ and an observation $O$ (an image of an object). If it thinks $D$ matches $O$, then it returns a positive conclusion; if it thinks $D$ does not match $O$, it returns a negative conclusion; if it is uncertain, it can ask a question. Every time it produces a reasoning $R$, motivating the decision. We map these interactions to the following $(R,S,Q)$ tuples, respectively: $(R,S=2,\text{None}), (R,S=0, \text{None}), (R,S=1,\text{Question})$. If there were no previous questions and answers, then $C = \emptyset$; otherwise, $C$ will contain the previous question-answer pairs. Therefore, each interaction maps directly to an entry of our dataset: $(D,O,R,S,Q,C)$.
In this way, we collected around 4,000 interactions from the training question-asking episode data and included them in the \textit{Train} split of \ourdataset. 
Finally, we combine those samples with the 24,000 samples from the previous phase to obtain the 28,000 samples of \ourdataset.

\begin{figure}
    \centering
    \includegraphics[width=0.49\linewidth]{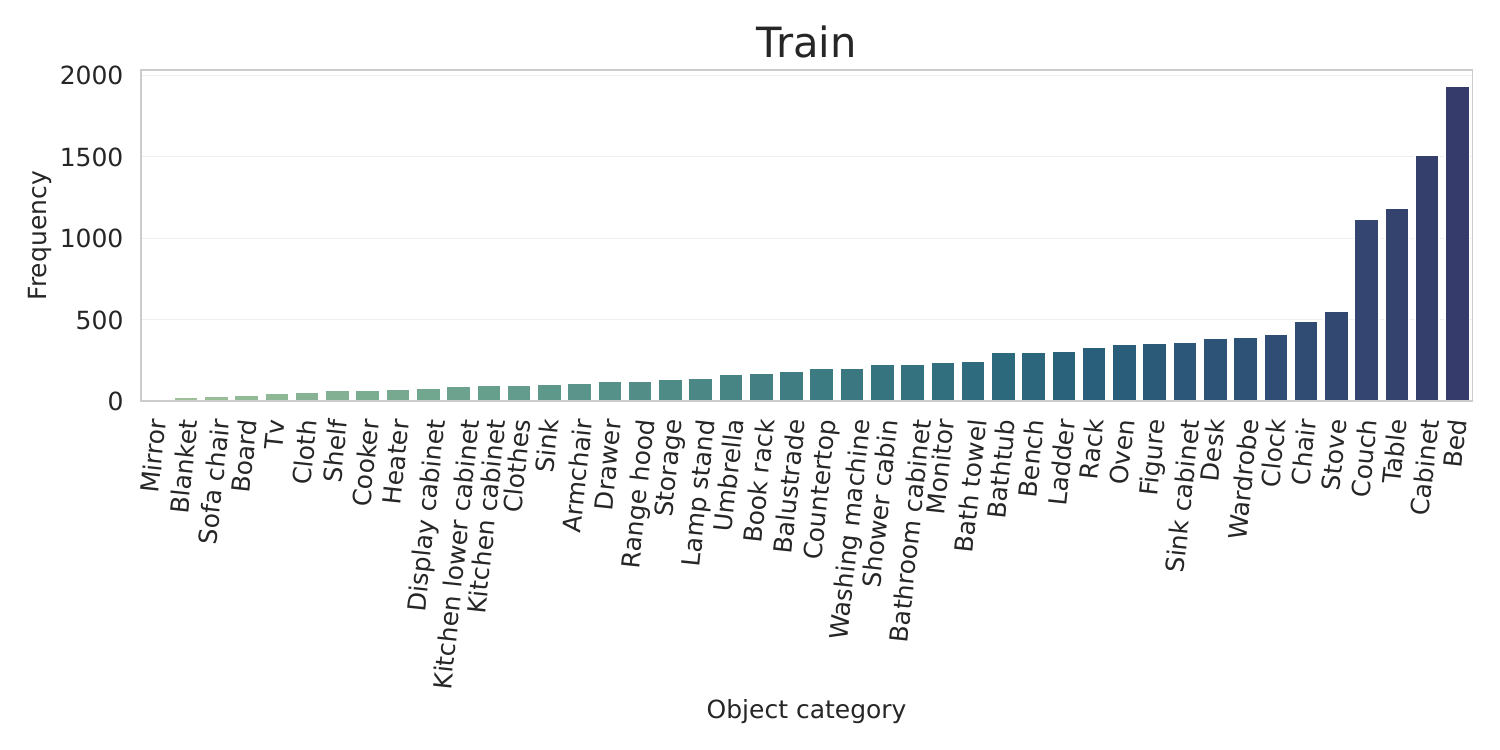}
    \includegraphics[width=0.49\linewidth]{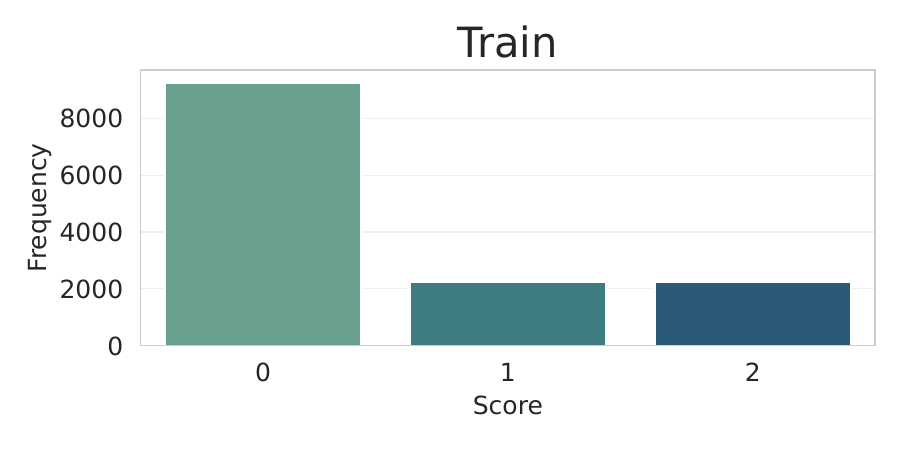}
    \includegraphics[width=0.49\linewidth]{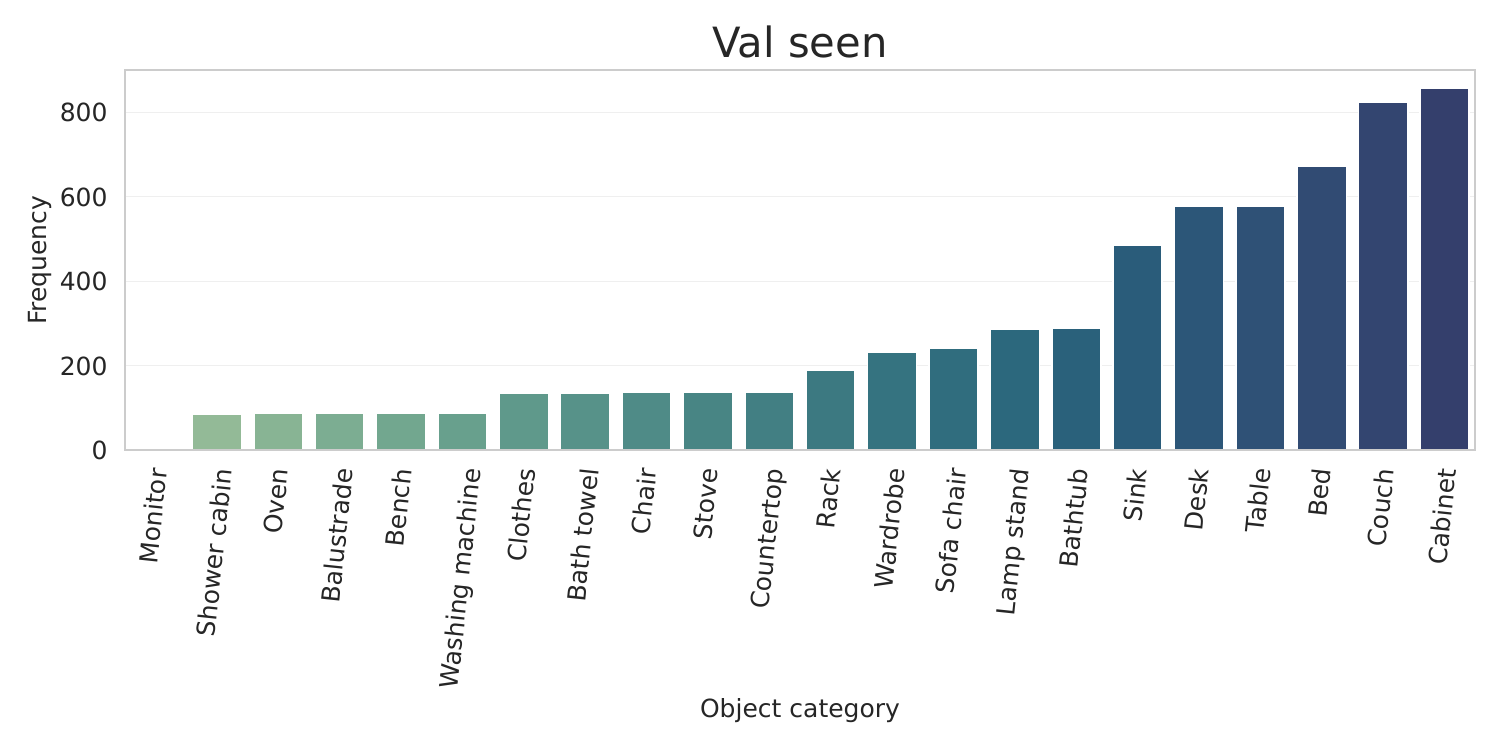}
    \includegraphics[width=0.49\linewidth]{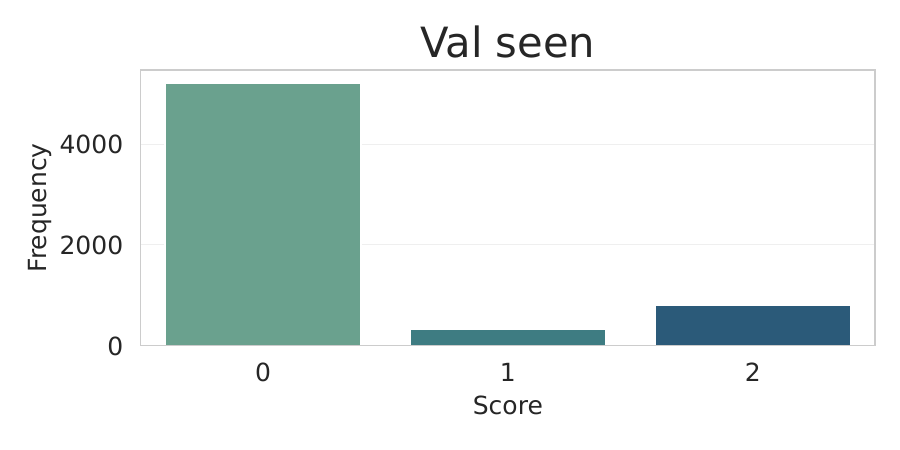}
    \includegraphics[width=0.49\linewidth]{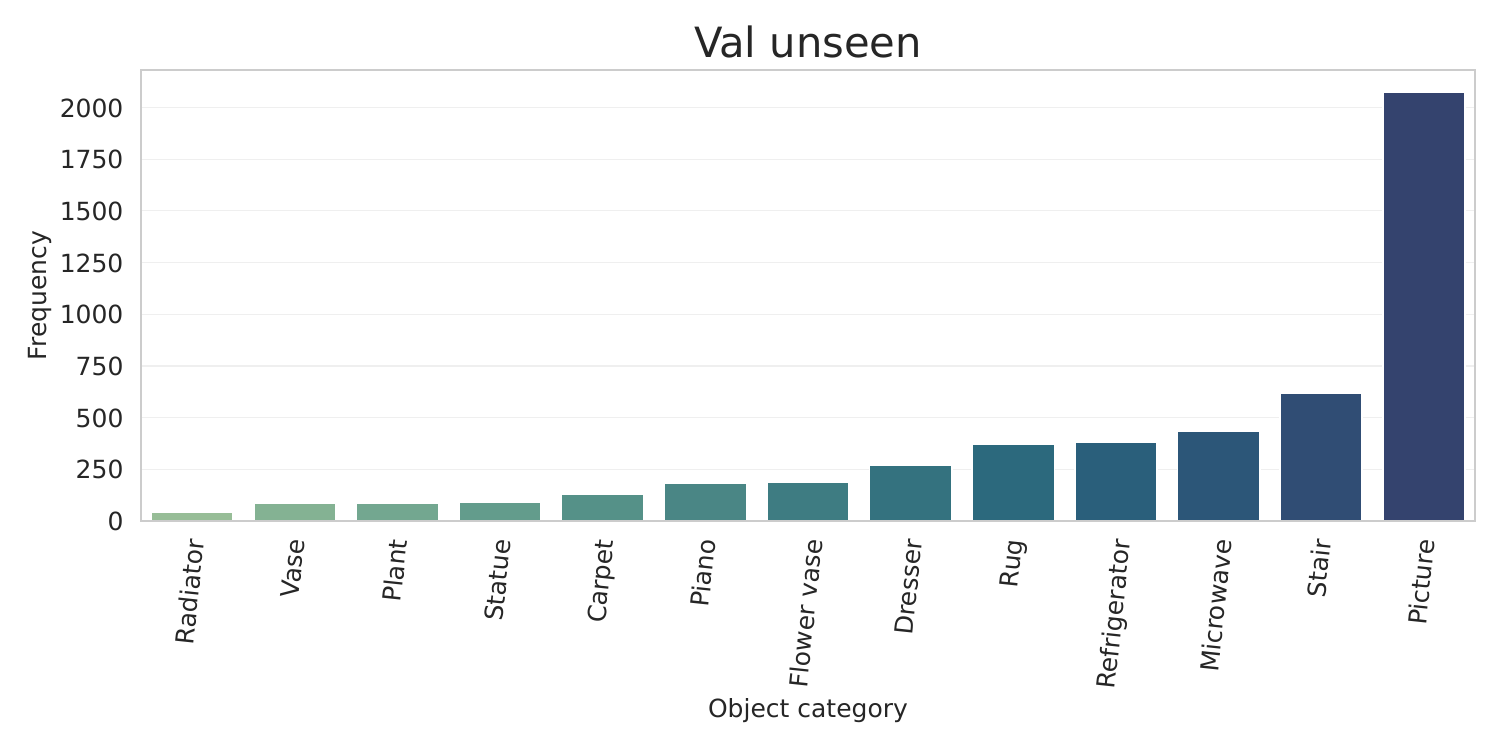}
    \includegraphics[width=0.49\linewidth]{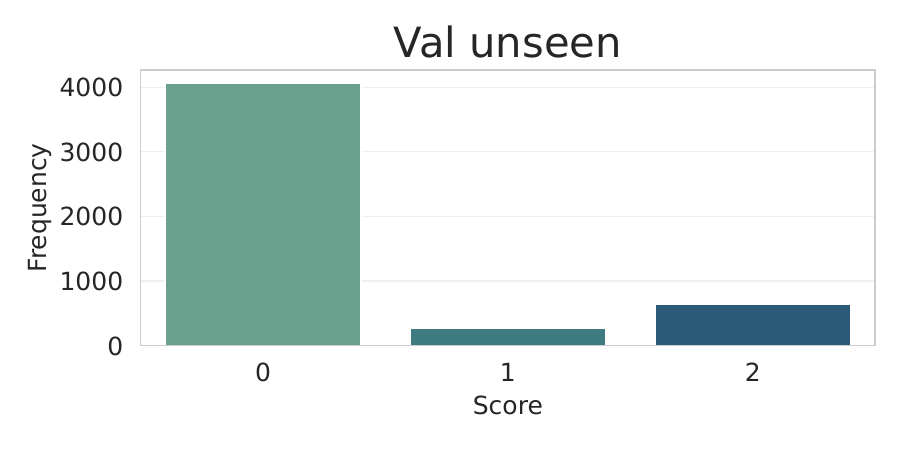}
    \caption{Distribution of the object categories in \ourdataset (left) and distribution of scores (right), divided by split. }
    \label{fig:dataset_distribution}
\end{figure}

\subsection{Dataset Statistics}
In \cref{fig:dataset_distribution}, we show information about \ourdataset. In particular, for each split (\textit{Train}, \textit{Val Seen} and \textit{Val Unseen}), we report (left) the object categories distribution, and (right) the distribution of scores.
We can see that the dataset contains, predominantly, $S=0$. This is expected as the images, in general, are quite different than each other, and therefore most image-description pairs do not match. The \textit{training} is more balanced, having as many $S=1$ (and therefore questions) as $S=2$. This is thanks to the second data collection phase, where we use rollouts of a strong \textit{questioner} on our question-asking protocol to produce high-quality questions and reasoning traces. For this reason, the training set contains more questions (and $S=1$) than the other two splits. 

\section{\ourmethod} \label{sec:appendix_method}

\subsection{Motivation of First Training Stage}\label{sec:appendix_training}

\begin{figure}
    \centering
    \includegraphics[width=0.8\linewidth]{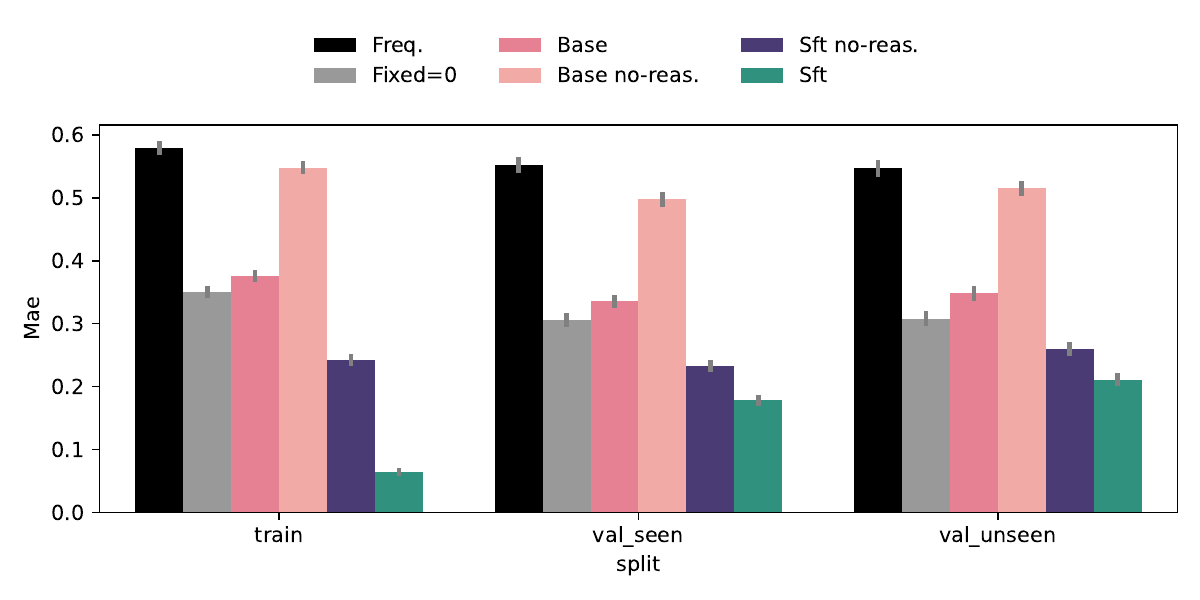}
    \caption{Ablation of different models' performance after the first training phase on \ourdataset. Performance is measured with the \textit{mean absolute error} between the output score and the ground truth score in the dataset. After this phase, a model can only return $R$ and $S$. }
    \label{fig:MAE_RQION}
\end{figure}

During the first training stage that we described in the main paper, \ourmethod learns the capability to perform reasoning $R$ and score $S$ for its uncertainty. 
We show here an experiment to motivate the need of this first training stage. 
In particular, \cref{fig:MAE_RQION} shows the results, in terms of \textit{mean absolute error} between the score $S$ from the model and the ground truth score, comparing our first-stage finetuned model as in the paper (\texttt{Sft}), a model finetuned using only $S$ (\texttt{Sft no-reas.}), the pre-trained base model (\texttt{Base}), the base model prompted to return only $S$ (\texttt{Base no-reas.}), a baseline always returning $S=0$ (\texttt{Fixed=0}), and a baseline picking scores based on their frequencies on the train set (\texttt{Freq.}).
We observe that returning both $R$ and $S$ (\texttt{Sft}) achieves the best performance across all splits of \ourdataset and models after fine-tuning. 
Because the dataset is unbalanced, returning always 0 is achieves a good performance, beating the untuned model. However, finetuning is always better, even in unseen splits. This is particularly meaningful for the val unseen split, where categories and images are completely new.

Having evaluated the strength of the first training stage against possible alternatives, we proceed with the second phase, where we train \ourmethod also on data points containing a question and a non-empty context. 
The results of the second training stage are reported in \cref{tab:results_qask} of the main paper. 

\subsection{Qualitative Examples}

\begin{figure}
    \centering
    \includegraphics[width=0.95\linewidth]{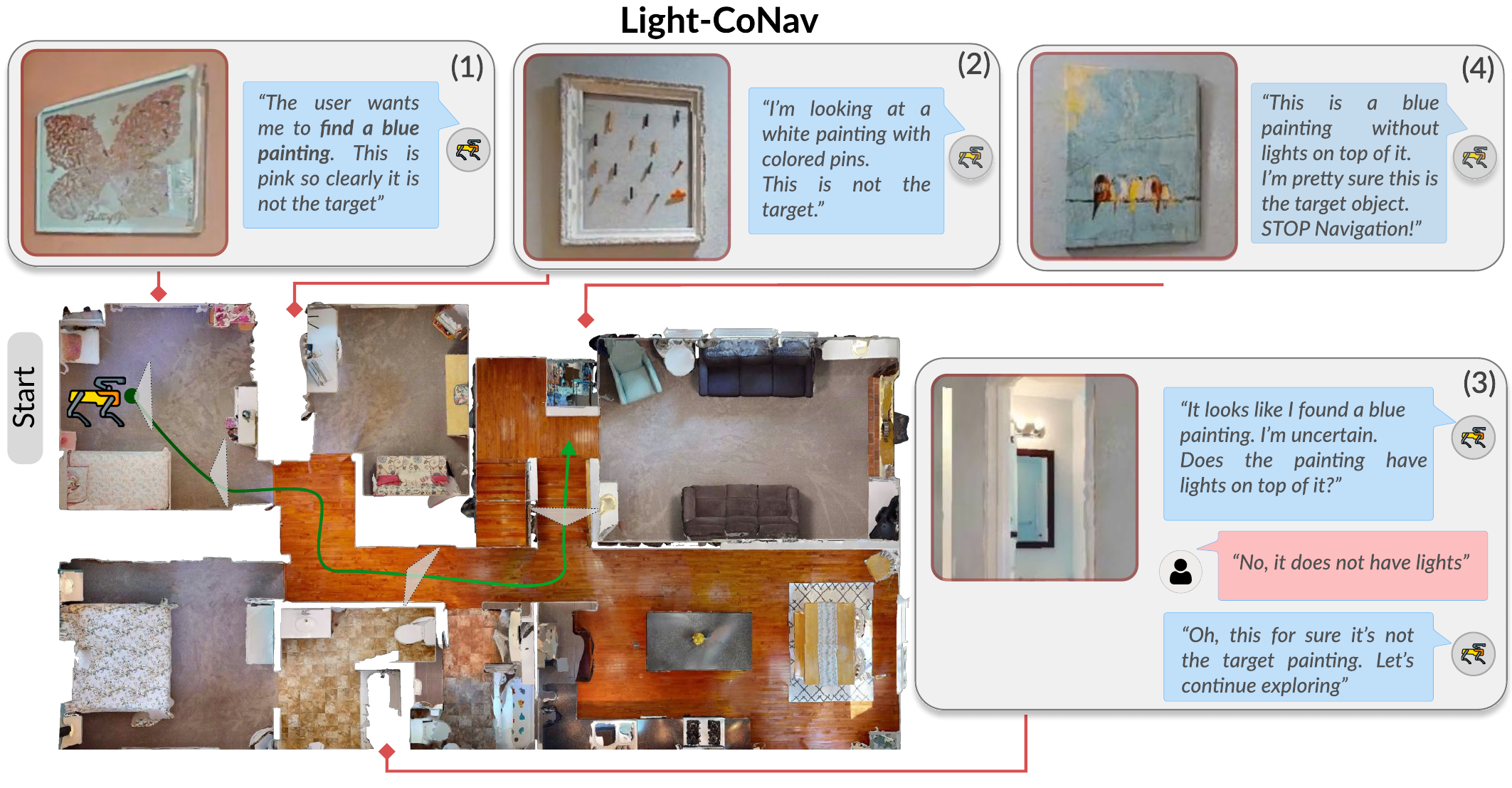}
    \caption{A CoIN episode. \ourmethod is looking for a `blue painting' ($D$). In the first two detections, $(1)$ and $(2)$, it correctly evaluates the observations, concluding that they are not the target object. $(3)$ is a difficult observation, as it looks like a painting. However, asking a question to the user disambiguates it. $(4)$ is the target object.}
    \label{fig:example_method}
\end{figure}

\cref{fig:example_method} reports a qualitative example of \ourmethod navigating a hard environment with different distractors. 
For visualization purposes, we report a short reasoning $R$, omit the score (which can be easily inferred from the conclusion), and append the question $Q$ after the reasoning. In practice, at each detection, the model returns $(R,S,Q)$ following the prompt in \cref{sec:appendix_prompts}.

The goal is to find a `blue painting' $(D)$. At the start (1), it immediately detects a painting; however, it correctly concludes that the painting is not the target blue painting, being pink ($S=0$). Next, after moving in the environment, it detects another painting (2). This time, too, it correctly concludes that this is not the target painting ($S=0$). The third observation is much harder: the agent mistakenly sees a mirror, reflecting blue light, as a possible painting, and stops the navigation to ask a question ($S=1$). However, the answer of the user clears its doubts by helping it localizing the true target painting, which is correctly detected at the end (4) ($S=2$).

In addition to this example, we also present more qualitative episodes in the last 60 seconds of the accompanying video, showcasing the interactive dynamics of \ourmethod.

\section{Prompts} \label{sec:appendix_prompts}

We describe here how \ourmethod is prompted to achieve the navigation goals. 
When a candidate object is detected from the current image, the observation $O$ and the initial task description $D$ are passed to \ourmethod using the following prompt (\texttt{USER}\_\texttt{TASK} is substituted with the description $D$ and the image is prepended to this text):
\begin{beautifulprompt}
\begin{lstlisting}[style=promptstyle]
You are a robot navigating an enclosed space. Your goal is to navigate to the correct object based on the user's commands. You were given the following task by the user '{USER_TASK}'. 

Currently, you are facing a scene represented by the given image. Reason about what you are seeing, comparing what you know about the task (given the user commands) and the given scene. 
For example, if the description is 'Black leather sofa near a lampstand' your reasoning process will be 'I'm currently observing a brown sofa which is different than black, making it unlikely to be the target sofa. Moreover, there is no lampstand near it, only a rug and a window' etc. 

If there are distortions or artifacts, do not focus on them, focus on the object at hand. At the end of the reasoning process, evaluate how well the provided image aligns with the user's task. 

Assign a confidence score based on the following scale: 
- 0: You are certain the image **DOES NOT** match the task. 
- 1: You are unsure whether the image matches the task or not. 
- 2: You are certain the image **DOES** match the task. 

Provide a concise reasoning (under 100 words). If you are unsure (score 1), ask an informative question to the user: this question should help you dispel any doubt about the current observation. Otherwise, just return ''. 
Use your reasoning output to ask the best question possible. Prefer asking questions at the start, rather than blindly coming to a conclusion, if the description is very short and does not include many details. The image might have distortions or digital artifacts: *NEVER* mention them in the question. Consider the information contained in the previous questions and answers, which are provided here.
\end{lstlisting}
\end{beautifulprompt}

\noindent Followed by either
\begin{beautifulprompt}
There are no previous questions or answers.
\end{beautifulprompt}
\noindent  or
\begin{beautifulprompt}
1. Question1 <|answer|>Answer1<|answer|>

2. Question2 <|answer|>Answer2<|answer|>

...
\end{beautifulprompt}
These two prompts are used, respectively, when the context is empty and when there are previous questions and answers in the context.

\noindent Lastly, the prompt includes the format to use:
\begin{beautifulprompt}
Strictly follow this output format: 
<motivation>Your reasoning here</motivation>
<score>0, 1, or 2</score>
<question>Your question or None (if score is not 1)</question>
\end{beautifulprompt}
Therefore, we pass $(D,O,C)$ to \ourmethod using this prompt and \ourmethod returns, in turn, the output $(R,S,Q)$.

\newpage

\end{document}